\documentclass[draft]{agujournal2019}
\usepackage{lmodern}      
\usepackage{amsmath}      
\usepackage{array}        
\usepackage{microtype}    
\usepackage{ragged2e}     
\usepackage{url} 
\usepackage{lineno}
\usepackage[inline]{trackchanges} 
\usepackage{soul}
\draftfalse

\journalname{}

\begin{document}
\justifying
%
%


\title{Seismology modeling agent: A smart assistant for geophysical researchers}

%
%




\authors{Yukun Ren\affil{1,2}, Siwei Yu\affil{1}, and Kai Chen\affil{2}, JianWei Ma\affil{1}}


\affiliation{1}{Harbin Institute of Technology, Harbin, China}
\affiliation{2}{Zhongguancun Academy, Beijing, China}




\correspondingauthor{Siwei Yu}{siweiyu@hit.edu.cn}
\correspondingauthor{Kai Chen}{kaichen@zgci.ac.cn}



\begin{keypoints}
\item We develop MCP servers for the SPECFEM seismic modeling suite, which can be integrated into modern AI agent systems.
\item The agentic workflow supports both fully automated and interactive human-in-the-loop seismic simulations.
\item The agent-driven framework lowers the barrier to seismic modeling, paving the way for future automated scientific research.
\end{keypoints}

%
%

%
%

\begin{abstract}
To address the steep learning curve and reliance on complex manual file editing and command-line operations in the traditional workflow of the mainstream open-source seismic wave simulation software SPECFEM, this paper proposes an intelligent, interactive workflow powered by Large Language Models (LLMs). We introduce the first Model Context Protocol (MCP) server suite for SPECFEM (supporting 2D, 3D Cartesian, and 3D Globe versions), which decomposes the entire simulation process into discrete, agent-executable tools spanning from parameter generation and mesh partitioning to solver execution and visualization. This approach enables a paradigm shift from file-driven to intent-driven conversational interactions. The framework supports both fully automated execution and human-in-the-loop collaboration, allowing researchers to guide simulation strategies in real time and retain scientific decision-making authority while significantly reducing tedious low-level operations. Validated through multiple case studies, the workflow operates seamlessly in both autonomous and interactive modes, yielding high-fidelity results consistent with standard baselines. As the first application of MCP technology to computational seismology, this study significantly lowers the entry barrier, enhances reproducibility, and offers a promising avenue for advancing computational geophysics toward AI-assisted and automated scientific research. The complete source code is available at \url{https://github.com/RenYukun1563/specfem-mcp}.
\end{abstract}

\section*{Plain Language Summary}
Simulating how seismic waves travel through the Earth is important for understanding earthquakes, 
imaging underground structures, and studying the planet’s interior. 
SPECFEM is a widely used open-source software package for these simulations, 
but its workflow can be difficult for many researchers. 
It requires editing many detailed text files and running long sequences of command-line instructions, 
which can be time-consuming and easy to get wrong.

In this work, we build a new way to use SPECFEM by connecting it to modern Artificial Intelligence (AI) systems. 
We create a set of Model Context Protocol (MCP) servers that allow human users to control SPECFEM with simple, 
conversational instructions, which are interpreted by a large language model that then operates SPECFEM on their behalf. 
Instead of manually setting up every file, researchers can tell the AI what they want to 
simulate, review intermediate results, and adjust the setup interactively. Researchers can freely choose between fully automated 
execution and interactive control for any simulation task.

Through several case studies, we demonstrate how this framework can be used in practice.
These examples suggest that the approach makes SPECFEM easier to use, reduces opportunities for human error, 
and provides a new pathway for AI-assisted and automated research in computational geophysics.

%
%

%


%
%
%
%

\section{Introduction}
Numerical simulation of seismic wave propagation serves as a cornerstone of modern geophysics, 
providing an indispensable tool for understanding earthquake physics, assessing seismic hazards, 
and exploring subsurface resources \cite{AkiRichards2002,Fichtner2010}. Against this backdrop, the open-source SPECFEM software 
suite (including SPECFEM2D, SPECFEM3D\_Cartesian, and SPECFEM3D\_Globe), developed based on the 
spectral-element method, has evolved over two decades of collaborative development and community 
building to become one of the most significant and powerful numerical simulation tools in the 
field today \cite{Patera1984,SerianiPriolo1994,KomatitschVilotte1998,KomatitschTromp1999,KomatitschTromp2002a,KomatitschTromp2002b,TrompKomatitschLiu2008,SPECFEMWeb2025}. Its strength lies not only in the high precision, computational efficiency, 
and large-scale parallel capabilities of its core solver, but also in the mature ecosystem of auxiliary 
tools built around it \cite{Carrington2008,Peter2011}. This toolchain covers the entire scientific lifecycle, 
from pre-processing (e.g., model building and meshing) to post-processing (e.g., data aggregation and visualization). 
Consequently, it has become the tool of choice for researchers and students worldwide for wave propagation modeling, 
underpinning countless cutting-edge scientific investigations from regional to global scales.

Despite the power of the SPECFEM suite and its ecosystem, its steep learning curve and complex, 
file-driven workflow present a formidable technical barrier for newcomers. Users are required not only to develop a deep understanding of numerous physical parameters, 
but also to manually reason about and manage the dependencies among a series of interconnected input files before executing cumbersome command-line operations.
Traditionally, the typical solution to such complexity has been the development of a Graphical User Interface (GUI). 
A well-designed GUI can indeed translate tedious command-line tasks and file editing into more intuitive clicks and 
form-filling. However, the limitations of GUIs are significant: they simplify operations but fail to understand intent. 
A GUI cannot autonomously translate a high-level user objective—such as ``simulate the ground motion response from an 
earthquake in a basin with undulating topography''—into the necessary sequence of specific configurations. 
The user must still manually translate this scientific intent into dozens of parameter settings. 
A new paradigm beyond the traditional GUI is therefore needed, one that ultimately allows users to focus on 
describing their objectives rather than on executing low-level software operations.

Recent breakthroughs in Large Language Models (LLMs) offer a revolutionary solution to this challenge. 
Artificial Intelligence (AI) agent systems built upon LLMs can comprehend high-level human intent and autonomously plan and execute multi-step tasks. 
However, for such an agent to effectively control a legacy scientific software package like SPECFEM, a critical gap must 
be bridged: the agent requires a standardized, machine-readable interface to programmatically interact with the software's 
functionalities, which are traditionally accessible only through manual file editing and command-line inputs. 
This is precisely the problem addressed by the Model Context Protocol (MCP)\cite{Anthropic2024MCP,MCPSpec2025,GoogleCloud2025MCP}. MCP specifies a standard for creating 
structured, discoverable toolsets that AI agents can understand and invoke. 

In this study, we directly address the aforementioned challenges by designing and implementing, 
for the first time, an MCP server suite for SPECFEM and integrating it with AI agents to create an intelligent, 
interactive geophysical simulation workflow driven by natural language. 
Our core contributions are as follows. 
(1) We build the first MCP-based service layer for the SPECFEM family by implementing three MCP servers for SPECFEM2D, SPECFEM3D\_Cartesian, and SPECFEM3D\_Globe, each exposing a suite of MCP tools that collectively turn these traditional codes into open computational engines amenable to programmatic control and intelligent automation.
(2) We implement a flexible, intent-driven workflow in which researchers can, for any simulation task, choose between fully automated execution and interactive human–AI control, instead of manually configuring every file and command, they tell the AI what they want to simulate, review intermediate results, and adjust the setup and simulation strategy interactively.
(3) We show that the new workflow helps lower the barrier to entry and improve scientific reproducibility by freeing researchers from tedious low-level operations while preserving researcher control and scientific judgment through its human-in-the-loop design.
(4) We argue that our methodology offers a general technical paradigm for modernizing other geophysical software, paving the way for integrating legacy codes into a broader ecosystem of AI-assisted and automated scientific research.

The complete code for this study is released as open source on GitHub to foster community development. The remainder of this 
paper is organized as follows. Section~\ref{sec:Background and Related Work} reviews the background of SPECFEM and the enabling technologies of LLM agents and MCP. 
Section~\ref{sec:AI-Driven Workflow} describes the overall system architecture. Section~\ref{sec:Case Studies} presents the five case studies that validate the effectiveness of our agent-driven 
workflow. Section~\ref{sec:Conclusion and Outlook} provides a conclusion and outlines future research directions.

\section{Background and Related Work}
\label{sec:Background and Related Work}

\subsection{Seismic Modeling with SPECFEM}
\label{subsec:specfem_background}

The SPECFEM software suite is a widely used open-source family of codes for simulating seismic
wave propagation across local, regional, and global scales using the spectral-element method (SEM).
In the SEM, the displacement field is represented by high-degree Lagrange polynomials evaluated at
Gauss--Lobatto--Legendre (GLL) interpolation points within quadrilateral (2D) or hexahedral (3D)
elements, which yield an approximately diagonal mass matrix when using Legendre bases. 
This structure enables
efficient explicit time integration while retaining the geometric flexibility of finite elements and
the high accuracy of spectral methods. For sufficiently smooth media and solutions, the SEM exhibits
spectral (often exponential) convergence with increasing polynomial degree, a property that has been
extensively analyzed in both the numerical analysis and seismological literature \cite{Patera1984,SerianiPriolo1994,KomatitschVilotte1998,KomatitschTromp1999,Fichtner2010}. 
These attributes render the SEM well suited 
for modeling the complex wave phenomena encountered in geophysical exploration and realistic Earth models.

The mathematical foundation of the software rests on the governing partial differential equations for wave propagation. 
In its simplest scalar form, the problem is described by the acoustic wave equation:
\begin{equation}
\frac{1}{v^2(\mathbf{x})} \frac{\partial^2 u(\mathbf{x},t)}{\partial t^2} - \nabla^2 u(\mathbf{x},t) = s(\mathbf{x},t),
\label{eq:scalar_wave}
\end{equation}
where $u(\mathbf{x},t)$ denotes a scalar wavefield, $v(\mathbf{x})$ is the
wave speed, and $s(\mathbf{x},t)$ represents a source term. 
While this scalar formulation captures basic wave kinematics, accurate Earth modeling generally requires solving the full vector elastodynamic system to account for complex solid mechanics. 
Consequently, the core solver of SPECFEM addresses the momentum equation in heterogeneous anisotropic media,
\begin{equation}
\rho(\mathbf{x}) \frac{\partial^2 u_i(\mathbf{x},t)}{\partial t^2} = \frac{\partial \sigma_{ij}(\mathbf{x},t)}{\partial x_j} + f_i(\mathbf{x},t), \qquad i = 1,2,3,
\label{eq:elastodynamic}
\end{equation}
coupled with the constitutive relations
\begin{equation}
\sigma_{ij}(\mathbf{x},t) = c_{ijkl}(\mathbf{x}) \varepsilon_{kl}(\mathbf{x},t), \qquad \varepsilon_{kl}(\mathbf{x},t) = \tfrac{1}{2} \left( \frac{\partial u_k}{\partial x_l} + \frac{\partial u_l}{\partial x_k} \right),
\label{eq:hooke}
\end{equation}
where $\rho$ is density, $\mathbf{u}$ is the displacement vector, $\boldsymbol{\sigma}$ and
$\boldsymbol{\varepsilon}$ are the Cauchy stress and strain tensors, $c_{ijkl}$ is the fourth-order
stiffness tensor, and $\mathbf{f}$ denotes body-force density. The SPECFEM codes apply the SEM to the
weak formulation of \eqref{eq:elastodynamic}--\eqref{eq:hooke} and their acoustic, viscoelastic, and
poroelastic variants\cite{AkiRichards2002,Fichtner2010}.

The SPECFEM family consists of several specialized packages tailored to different modeling scenarios. \texttt{SPECFEM1D}
serves primarily as an educational code for one-dimensional heterogeneous media, illustrating the
fundamental principles of SEM discretization and time integration. \texttt{SPECFEM2D} simulates forward and
adjoint wave propagation in 2D Cartesian and axisymmetric (2.5D) configurations for acoustic, elastic,
viscoelastic, and poroelastic media, and implements convolutional perfectly matched layers (C-PMLs) \cite{KomatitschMartin2007} to
suppress artificial reflections at truncated boundaries. \texttt{SPECFEM3D\_Cartesian} extends these
capabilities to 3D regional and local-scale simulations on conforming hexahedral meshes produced by
external meshing tools such as CUBIT or Gmsh, supporting acoustic, (an)isotropic elastic,
poroelastic, and coupled solid--fluid models as well as adjoint tomography and full-waveform inversion. \texttt{SPECFEM3D\_GLOBE} targets global and
regional wave propagation in spherical geometries, incorporating 3D heterogeneity, ellipticity,
topography and bathymetry, ocean loading, rotation, and self-gravitation on a cubed-sphere mesh, and
has been used to perform high-frequency global simulations on modern supercomputers \cite{SPECFEMWeb2025,SPECFEM3DCartesianManual,KomatitschVilotte1998,KomatitschTromp1999,KomatitschTromp2002a,KomatitschTromp2002b,TrompKomatitschLiu2008}.

The SEM addresses several challenges that are central to accurate seismic wave modeling. First, it
represents complex free-surface topography without stair-stepping artifacts. Second, it treats
solid--fluid interfaces in marine and crustal environments with high fidelity. Third, it flexibly
incorporates strong lateral heterogeneity, anisotropy, and anelastic attenuation parameterized by
quality factors. 
Finally, SEM discretizations
admit efficient domain decomposition and exhibit excellent strong and weak scaling on large
distributed-memory systems, making SPECFEM a reference code in computational seismology and
seismic hazard analysis\cite{Carrington2008,Peter2011}.

Despite these strengths, the traditional SPECFEM workflow remains technically demanding. According to
the official manuals and numerous application studies, a typical
simulation requires: (i) defining the Earth model and generating a high-quality mesh with external
software; (ii) configuring dozens to hundreds of parameters across multiple text-based input files
(\texttt{Par\_file}, \texttt{CMTSOLUTION}, \texttt{STATIONS}...); (iii)
invoking a sequence of separate executables for meshing (\texttt{xmeshfem*}), database generation
(\texttt{xgenerate\_databases}), and time stepping (\texttt{xspecfem*}); (iv) managing parallel runs
with MPI across heterogeneous HPC environments; and (v) post-processing and visualizing large
seismogram and field datasets. This file-centric, command-line-driven paradigm---often orchestrated
through handwritten shell scripts and job submission templates---creates a steep learning curve and
substantial potential for configuration errors. 
This motivates the search for higher-level, intent-driven interfaces that preserve the numerical performance 
of SPECFEM while substantially reducing operational complexity\cite{SPECFEMWeb2025,SPECFEM3DCartesianManual}.

\subsection{LLM-based Autonomous Agents}
\label{subsec:llm_agents}

LLMs have moved beyond passive text generation and are increasingly used as
autonomous agents capable of perceiving their environment, planning, and executing complex tasks.
Compared with standard conversational models, typical agent architectures augment the LLM backbone
with three key components: (i) a \emph{planning} module that decomposes high-level goals into
sequences of executable sub-tasks; (ii) a \emph{memory} mechanism that maintains and updates
relevant context over extended time horizons; and (iii) a \emph{tool-use} interface that allows the
model to call external APIs, run code, or operate existing software systems. 
This paradigm shift enables LLMs to operate not merely as text generators but as goal-directed problem-solving systems that 
map natural-language instructions to structured sequences of decisions and actions.
Several recent surveys provide systematic overviews of LLM-based autonomous agents \cite{durante2024agent,xi2025rise,cheng2024exploring}.

A number of representative design paradigms have emerged for LLM agents. The ReAct framework \cite{Yao2023ReAct}
interleaves chain-of-thought reasoning with actions: at each step, the model generates both an
explicit reasoning trace and an action to interact with tools or environments, and then conditions
subsequent decisions on the observed outcomes. This simple pattern has become a
foundational template for many later tool-using and multi-agent systems. Plan-and-Solve proposes a
two-stage prompting strategy in which the model first produces a global solution plan and then
executes the plan step by step, leading to substantial gains on multi-step mathematical and symbolic
reasoning tasks in a zero-shot setting\cite{Wang2023PlanSolve}. Complementary to these methods,
Executable Code Actions / CodeAct-style approaches represent an agent’s actions as executable code
snippets (e.g., Python), enabling flexible control- and data-flow over multiple tool calls while
using runtime errors and outputs for self-debugging and refinement; such designs have shown improved
efficiency and success rates on complex tool-use benchmarks compared with traditional JSON-based
tool calling\cite{Chen2025ScienceAgentBench}. More recently, DeepAgent \cite{li2025deepagent} system pushes
towards higher degrees of autonomy: within a single long reasoning trajectory, the agent discovers
relevant tools, plans and executes sequences of calls, and compresses multi-turn interactions into
structured episodic, working, and tool memories, while reinforcement learning is used to train the
overall decision policy in large-scale simulated API environments.

There is mounting evidence that LLM-based agents can support intricate workflows and contribute to
scientific discovery. On the evaluation side, new benchmarks aim to approximate real research
processes. For example, PaperBench evaluates agents by asking them to
reproduce full machine learning papers: starting from the publication text, the agent must implement
code, design and run experiments, and reproduce key results, with evaluation combining rubric-based
LLM judgments and human review\cite{OpenAI2025PaperBench}. Other efforts, such as ScienceAgentBench, decompose
data-driven scientific discovery into stages of data cleaning, feature engineering, experimental
design, and result analysis, measuring an agent’s ability to coordinate end-to-end ``data--code--conclusion''
pipelines\cite{Chen2025ScienceAgentBench}. At the same time, top-tier journals have begun to publish
closed-loop scientific agent systems that interface with the physical world. A notable example is
the ``virtual lab'' framework reported in \emph{Nature}, where an LLM-based ``principal
investigator'' coordinates multiple specialist agents to design nanobody sequences against emerging
SARS-CoV-2 variants, orchestrating protein language models, structure prediction tools, and
molecular modeling software, and ultimately proposing antibody candidates that are validated
experimentally to exhibit strong binding\cite{Swanson2025VirtualLab}. 
In geophysics and the broader geosciences, LLM-based agents are at a comparatively early stage, but
several pioneering works illustrate their potential. A recent article \cite{kanfar2025intelligent}
demonstrates how generative LLMs can be integrated with industrial software to form ``intelligent
seismic workflows'': geophysicists express goals and constraints in natural language, and an agent
proposes processing flows, configures parameters, launches existing algorithms, and performs
quality-control on intermediate results, thereby raising the level of abstraction for seismic data
processing and interpretation.

In summary, LLM-based autonomous agents are evolving from simple conversational assistants into
systems capable of coordinating complex scientific and engineering workflows, often in the form of
collaborating agent populations. In the general AI domain, frameworks such as ReAct,
Plan-and-Solve, CodeAct, and DeepAgent have laid the foundations for tool use, long-horizon
planning, multi-step reasoning, and reinforcement learning-based training paradigms. In scientific
discovery, benchmarks like PaperBench and ScienceAgentBench, together with multi-agent systems
reported in leading journals, demonstrate increasingly end-to-end capabilities in realistic
research scenarios. Against this backdrop, the present work focuses
on building toolized interfaces around the spectral-element seismic modeling software SPECFEM and
embedding them into a general LLM agent framework, in order to explore the deployment of LLM-based
agents in rigorous physics-based simulation workflows.

\subsection{Tool Use and the Model Context Protocol}
A central requirement for LLM-based agents is a standardized mechanism for discovering and invoking
external tools and data sources. Historically, each AI application implemented its own function-calling
API or plugin system, leading to fragmented, vendor-specific ecosystems.
The Model Context Protocol, introduced by Anthropic on November 2024, addresses this challenge by providing an
open-source, vendor-neutral standard for connecting AI applications to external tools and data\cite{Anthropic2024MCP,GoogleCloud2025MCP}.
The protocol's design draws inspiration from the Language
Server Protocol \cite{MCPSpec2025}, which successfully standardized interactions between programming languages
and development tools, and implements its message-passing semantics using JSON-RPC 2.0 over standard
transport mechanisms including standard input/output (stdio) and HTTP with Server-Sent Events (SSE).

\begin{figure}[htbp]
  \centering
  \includegraphics[width=\textwidth]{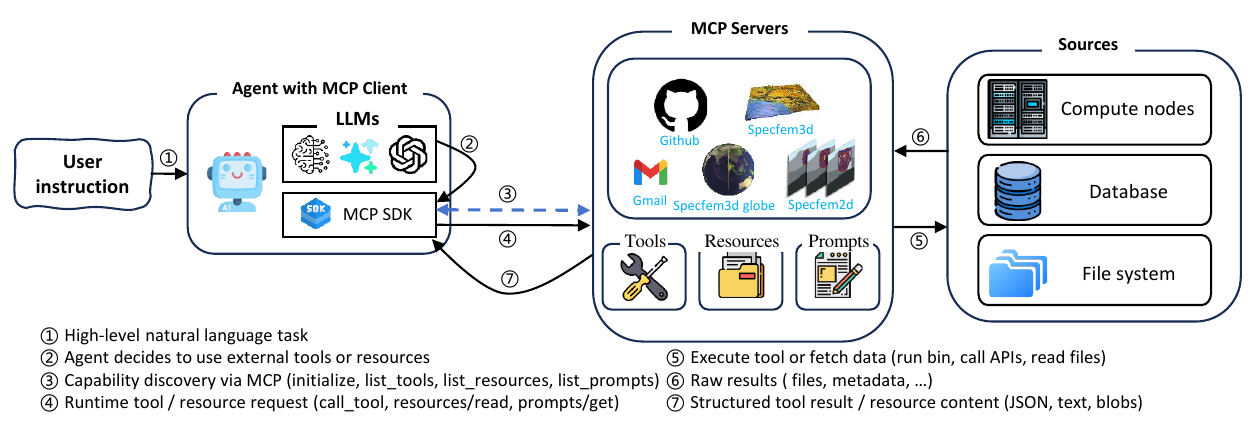}
  \caption{Overview of an MCP-based agent workflow. A user issues a high-level natural-language
  instruction to an agent that embeds an MCP client SDK. The LLM decides when external
  tools or resources are needed and asks the MCP client to discover and invoke MCP servers
  via the Model Context Protocol. Each server follows a simple client–server pattern: the
  client opens a connection and sends JSON-RPC requests, the server invokes the corresponding
  execution logic on underlying sources (e.g., compute nodes, databases, file systems), and
  returns structured results (JSON, text, blobs) that are injected into the LLM context}
  \label{fig:mcp_architecture}
\end{figure}

MCP's architecture follows a client-server model (Figure~\ref{fig:mcp_architecture}), where MCP clients, typically embedded within
LLM-powered applications such as AI assistants or integrated development environments, initiate
connections to MCP servers that expose specific functionalities. Each server advertises a catalog of
available tools through a \texttt{list\_tools} operation, with each tool accompanied by a
comprehensive JSON schema that specifies its name, description, input parameters, and expected
output format. When an LLM agent determines that a particular tool is needed to fulfill a user
request, the client formulates a \texttt{call\_tool} message containing the tool name and properly
formatted arguments, transmits it to the server, which then executes the underlying function and
returns results in a standardized structure. This request-response cycle enables agents to
seamlessly interact with heterogeneous systems without requiring custom integration code for each
data source\cite{MCPSpec2025}.

The protocol defines three core primitives that structure agent-environment interactions.
\textit{Tools} represent executable functions that the LLM can invoke autonomously, such as querying
databases, performing calculations, or controlling external software. The agent decides when to use
tools based on its reasoning about task requirements. \textit{Resources} provide structured data or
context that enriches the model's understanding, including document excerpts, code snippets, or
configuration files. Resources are typically managed by the host application rather than directly by
the LLM, allowing fine-grained control over context injection. \textit{Prompts} define reusable
templates or instruction sequences that guide the model's behavior in specific scenarios, enabling
consistent handling of recurring tasks. Beyond these primitives, MCP incorporates additional
capabilities including \textit{sampling}, which allows servers to request LLM completions from
clients for multi-step reasoning, and \textit{roots}, which define entry points into the host's file
system with appropriate permission boundaries\cite{MCPSpec2025}.

Since its release, MCP has garnered significant adoption across the AI ecosystem. A diverse array of LLM 
providers, host applications, and developer tools have rapidly implemented support for the protocol\cite{GoogleCloud2025MCP}. 
This broad adoption has catalyzed a vibrant community ecosystem populated with extensive libraries of 
ready-to-use servers and integrations. This architectural shift promises to foster a collaborative ecosystem 
where the community contributes reusable servers, dramatically accelerating the development of context-aware AI 
applications across scientific and commercial domains.

\section{AI-Driven Workflow}
\label{sec:AI-Driven Workflow}
\begin{figure}[htbp]
    \centering
    \includegraphics[width=\textwidth]{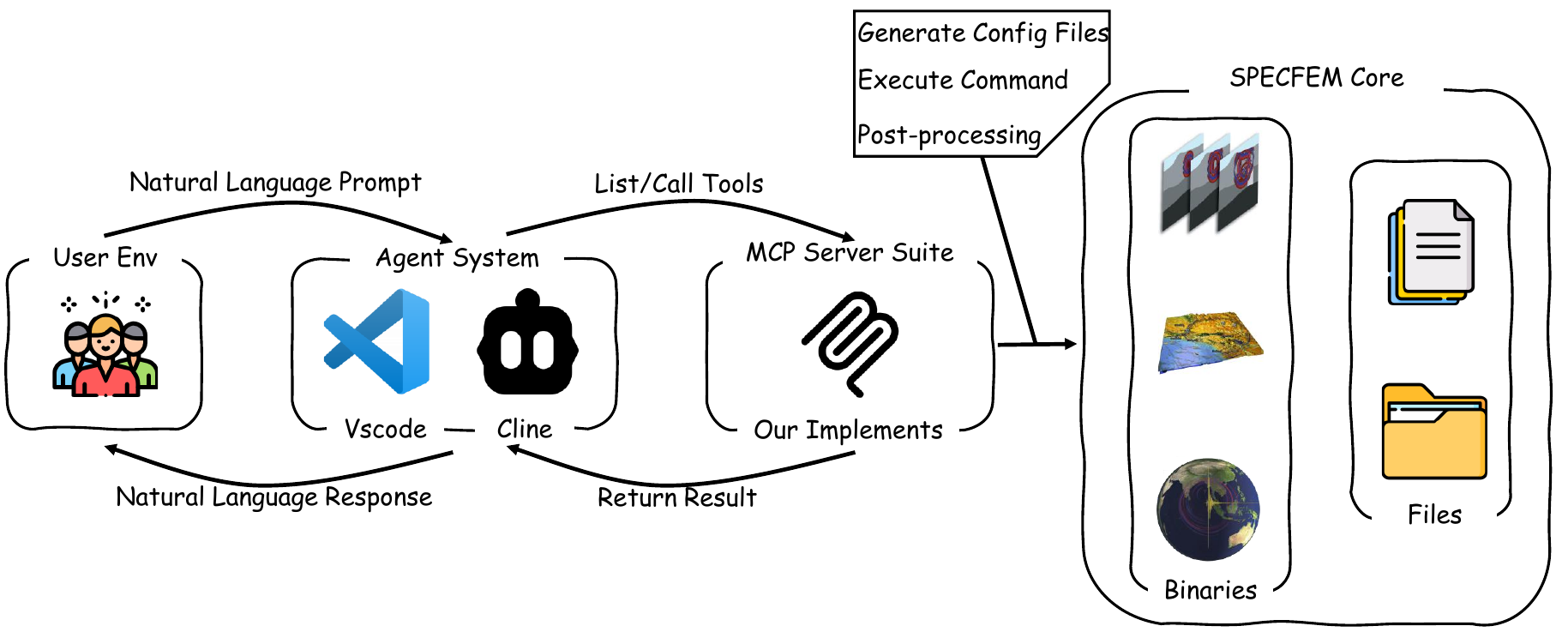}
    \caption{System architecture diagram showing the interaction between the user, the LLM agent
    (the \texttt{cline} plugin), the SPECFEM MCP servers, and the core SPECFEM executables. The
    \texttt{cline} plugin encompasses both the user-facing UI and the backend agent logic.}
    \label{fig:architecture}
\end{figure}

In this section, we detail the architecture of our AI-driven workflow and the implementation of SPECFEM MCP servers. 
The overall system architecture is illustrated in Figure~\ref{fig:architecture}.

The workflow begins with a user's prompt. The \texttt{cline} plugin in Visual Studio Code captures
this prompt and passes it to its backend agent logic, which is aware of the available MCP tools and
orchestrates the simulation pipeline\cite{ClineVSCode2024,ClineWebsite2025,MCPClients2025}. For instance, a request to ``simulate an earthquake'' may
trigger a sequence of tool calls: first to generate the necessary input files (e.g., \texttt{interfaces.dat}, \texttt{Par\_file}, \texttt{STATIONS}...), followed by
tools that run the mesher and then the solver. The results---such as confirmation of file
generation, simulation completion status, and paths to output visualizations---are returned to the
agent backend, which synthesizes a natural-language report and presents it to the user, thereby
completing the intent-driven workflow. At any point, the user can intervene to adjust parameters or
inspect intermediate results, so the workflow naturally supports both automated and
human-in-the-loop operation.

\textbf{\textit{SPECFEM software suite:}} This is the underlying high-performance scientific
software that performs the numerical simulations. Our workflow does not modify the core SPECFEM
code. Instead, the MCP servers interact with SPECFEM in the same way as a traditional command-line
user: they prepare the required input files in a working directory such as \texttt{DATA/}, invoke
the appropriate compiled binaries in \texttt{bin/}, and then read the output files written to
\texttt{OUTPUT\_FILES/} for visualization or further analysis.

\textbf{\textit{SPECFEM MCP servers:}} This layer forms the core of our technical contribution. We
develop three independent MCP servers using an asynchronous architecture, one for each SPECFEM
component (SPECFEM2D, SPECFEM3D\_Cartesian, SPECFEM3D\_Globe). These servers act as middleware,
translating the agent's programmatic requests into file- and process-level operations that the
SPECFEM codes can execute. The key tools implemented for each server are summarized in
Table~\ref{tab:specfem2d_tools}, Table~\ref{tab:specfem3d_tools}, and
Table~\ref{tab:specfem3d_globe_tools}.

Each MCP server dynamically loads tool modules from its \texttt{tools/} directory. Every tool module
implements two key functions: \texttt{get\_tool}, which exposes the tool's definition and JSON
schema to the agent, and \texttt{handle}, which contains the logic to execute the tool. This logic
often uses Jinja2 templates to generate configuration files (e.g., \texttt{Par\_file},
\texttt{STATIONS}) or wraps SPECFEM's compiled Fortran executables (e.g., \texttt{xmeshfem2D},
\texttt{xspecfem3D}) in a subprocess while managing file paths and MPI parallelization.

\textbf{\textit{Agent with MCP client:}} The core of the user interaction and workflow orchestration
is the \texttt{cline} plugin, a comprehensive agent system integrated into Visual Studio Code \cite{ClineVSCode2024,ClineWebsite2025}. It
consists of two parts:
\begin{itemize}
    \item \textbf{Frontend (chat UI):} The user-facing component inside the VS Code editor. It
    provides a chat-like interface where researchers can express scientific objectives in natural
    language, upload auxiliary files (such as pre-existing mesh data), and receive feedback,
    generated files, and visualizations directly within their IDE.

    \item \textbf{Backend (agent logic):} The backend is the ``brain'' of the agent. Upon receiving a
    prompt from the frontend, its LLM-based logic performs several functions: (1) \textbf{Intent
    recognition} to identify the scientific goal; (2) \textbf{Tool discovery} by querying the MCP
    servers for their capabilities via \texttt{list\_tools} requests; (3) \textbf{Planning} a
    sequence of steps to achieve the goal; (4) \textbf{Tool execution} by formulating valid JSON
    objects and sending \texttt{call\_tool} requests to the MCP servers; and (5) \textbf{Response
    synthesis} to process the results and formulate a coherent natural-language response for the
    user.
\end{itemize}

\begin{table}[htbp]
\centering
\caption{Tools Provided by the SPECFEM2D MCP Server}
\label{tab:specfem2d_tools}
\renewcommand{\arraystretch}{1.3}
\begin{tabular}{l m{0.65\textwidth}}
\hline
\textbf{Tool Name Suffix} & \textbf{Function} \\
\hline
\texttt{generate\_par\_file} & Generates the main configuration file for SPECFEM2D simulations. \\
\texttt{generate\_source\_file} & Creates the seismic source configuration file. \\
\texttt{generate\_stations\_file} & Creates the receiver station configuration file. \\
\texttt{generate\_interfaces\_file} & Creates the geological layer interface definition file for the internal mesher. \\
\texttt{run\_mesher} & Executes the 2D mesher. \\
\texttt{run\_solver} & Executes the 2D seismic wave propagation solver. \\
\texttt{visualize} & Visualizes simulation results. \\
\hline
\end{tabular}
\end{table}

\begin{table}[htbp]
\centering
\caption{Tools Provided by the SPECFEM3D\_Cartesian MCP Server}
\label{tab:specfem3d_tools}
\renewcommand{\arraystretch}{1.3}
\begin{tabular}{l m{0.65\textwidth}}
\hline
\textbf{Tool Name Suffix} & \textbf{Function} \\
\hline
\texttt{generate\_par\_file} & Generates the main configuration file for SPECFEM3D simulations. \\
\texttt{generate\_cmtsolution\_file} & Creates a CMTSOLUTION file for moment tensor sources. \\
\texttt{generate\_forcesolution\_file} & Creates a FORCESOLUTION file for point force sources. \\
\texttt{generate\_stations\_file} & Creates the receiver station configuration file. \\
\texttt{run\_mesh\_generator} & Executes the internal mesh generator. \\
\texttt{decompose\_mesh} & Decomposes mesh for MPI parallel processing. \\
\texttt{generate\_databases} & Generates model databases based on the decomposed mesh. \\
\texttt{run\_solver} & Executes the 3D seismic wave propagation solver.\\
\texttt{visualize} & Visualizes simulation results. \\
\hline
\end{tabular}
\end{table}

\begin{table}[htbp]
\centering
\caption{Tools Provided by the SPECFEM3D\_Globe MCP Server}
\label{tab:specfem3d_globe_tools}
\renewcommand{\arraystretch}{1.3}
\begin{tabular}{l m{0.65\textwidth}}
\hline
\textbf{Tool Name Suffix} & \textbf{Function} \\
\hline
\texttt{generate\_par\_file} & Generates the main configuration file for global simulations. \\
\texttt{generate\_cmtsolution\_file} & Creates a CMTSOLUTION file for moment tensor sources.  \\
\texttt{generate\_forcesolution\_file} & Creates a FORCESOLUTION file for point force sources. \\
\texttt{generate\_stations\_file} & Creates the receiver station configuration file. \\
\texttt{run\_mesher} & Executes the global mesh generator. \\
\texttt{run\_solver} & Executes the global solver. \\
\texttt{visualize} & Visualizes simulation results. \\
\hline
\end{tabular}
\end{table}

\section{Case Studies}
\label{sec:Case Studies}
\begin{figure}[htbp]
    \centering
    \includegraphics[width=\columnwidth]{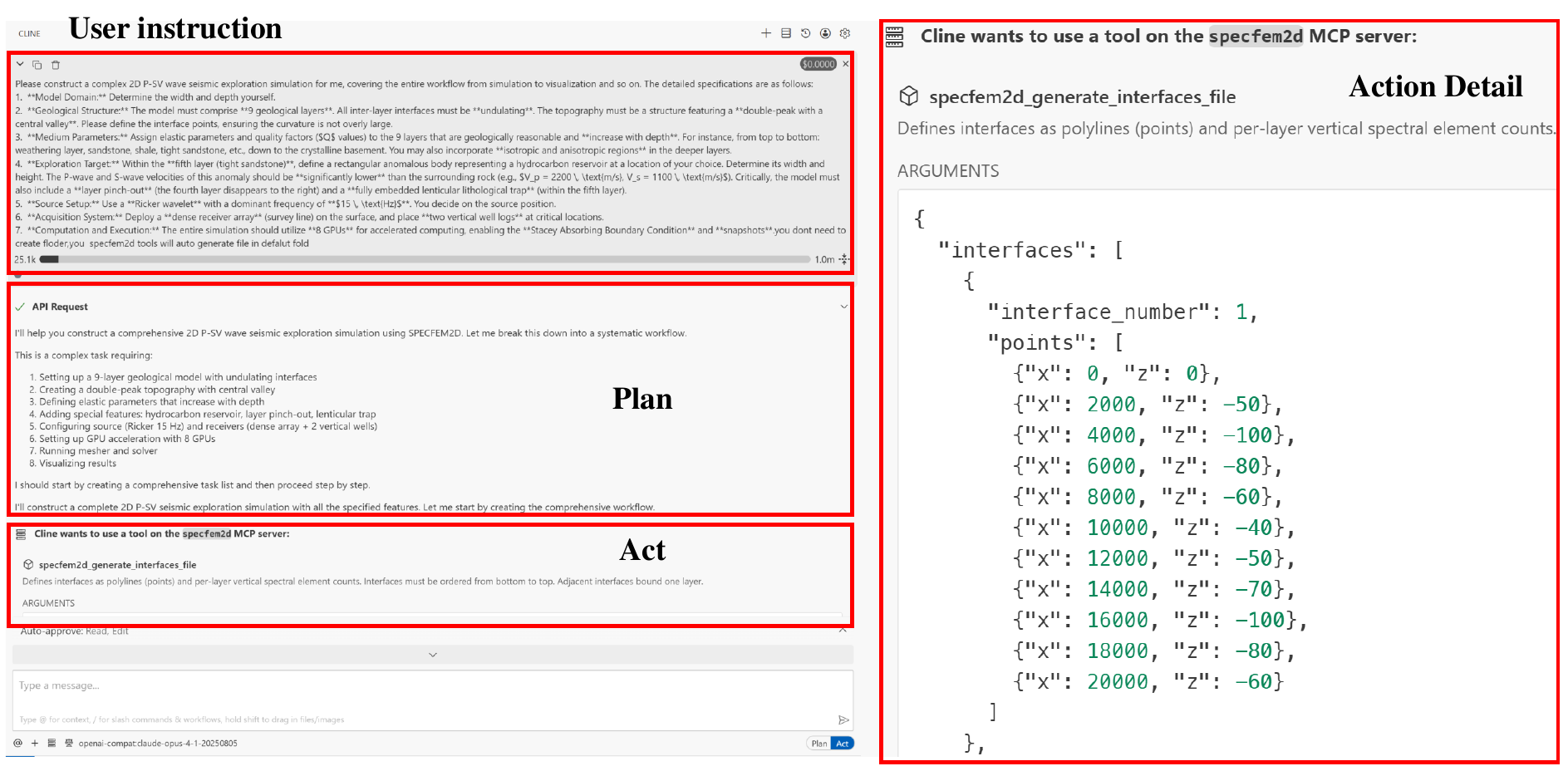}
    \caption{Actual execution of the \texttt{cline} agent operating under user-provided
    natural-language instructions.}
    \label{fig:cline_run}
\end{figure}

We demonstrate the capabilities of our system through five case studies, each highlighting a
different aspect of the LLM-driven workflow. We provide a practical example of the
\texttt{cline} agent with the SPECFEM MCP servers (Figure~\ref{fig:cline_run}),
illustrating how natural-language instructions are translated into simulation operations. To further
assess the generality of our workflow, we test it with several state-of-the-art LLMs, including
Google’s Gemini~2.5~Pro and Anthropic’s Claude~Opus~4.1, confirming that the MCP-based SPECFEM interface is compatible with
agents built on different model architectures.

\subsection{Case 1: Agent-Generated Teaching-Level Experiment Demonstrating the Seismic Lens Effect}

\subsubsection{Objectives and Setup}

The objective of this case study is to design a simple, teaching-level numerical experiment that 
illustrates the seismic ``lens effect''—a focusing phenomenon in which complex subsurface geometry 
accelerates and concentrates up-going wave energy, analogous to optical lensing. To reveal this 
effect clearly, we construct a comparative pair of 2D models that share identical material 
properties, source parameters, and receiver layouts, differing only in their subsurface geometry 
(Figure~\ref{fig:case1_snapshots}).

\begin{itemize}
    \item \textbf{Model A (Flat-Layer Control):} A horizontally stratified reference model with perfectly flat interfaces. It shares the identical vertical material sequence as the edges of Model B, serving as a baseline to isolate the effects of lateral heterogeneity.
    \item \textbf{Model B (Buried-Ridge Model):} A model featuring a composite high-velocity basement ridge composed of a core ($V_p = 5500$ m/s) and a high-velocity cap ($V_p = 6200$ m/s). The structure rises steeply from a depth of $-8$ km to a peak elevation of $-2.8$ km and is overlain by stratified low-velocity sediments ($V_p$: 1600--2800 m/s). This geometry acts as a high-velocity lens, accelerating the central wavefront and focusing up-going wave energy.
\end{itemize}

Both models employ a 2~Hz Ricker wavelet point source located at $(x = 10{,}000~\text{m},\, z = -12{,}000~\text{m})$. The resulting wavefields are recorded by a dense surface array of 201 receivers uniformly spaced at 250~m intervals, spanning the entire model width ($x=0$ to $50$~km).

\subsubsection{Agent-Driven Interactive Workflow}
The user provides a high-level scientific intent: 
``Use SPECFEM2D to design and execute a comparative experiment demonstrating the seismic focusing effect.
Construct two models--one with flat topography and one with a buried anticline--maintaining identical
material properties while varying only the subsurface geometry. Perform the complete forward simulation
workflow and visualize the resulting seismograms for both.'' 

Upon receiving this instruction, the agent initiates a ``Think--Plan--Execute'' procedure through the SPECFEM2D MCP server. Following the initial planning phase, the user refines specific details through a multi-turn dialogue with the agent. It then automatically generates the required input files (\texttt{interfaces.dat}, \texttt{Par\_file}, \texttt{SOURCES} and \texttt{STATIONS}), executes the meshing and forward solvers, and organizes the outputs for visualization. When necessary, the agent requests confirmation from the user; furthermore, under user supervision, it autonomously diagnoses execution errors and rectifies erroneous parameters.

This contrasts with traditional SPECFEM usage, in which users must manually create and validate several configuration files, execute multiple command-line programs in sequence, and locate the outputs for post-processing. By delegating these steps to the agent—while still enabling interactive refinement—the workflow becomes substantially more efficient and less error-prone.

\subsubsection{Results and Discussion}
The simulation results clearly illustrate the seismic lens effect, visible both in the evolution of 
the wavefield and in the surface seismograms.

\begin{figure}[htbp]
    \centering
    \includegraphics[width=\columnwidth]{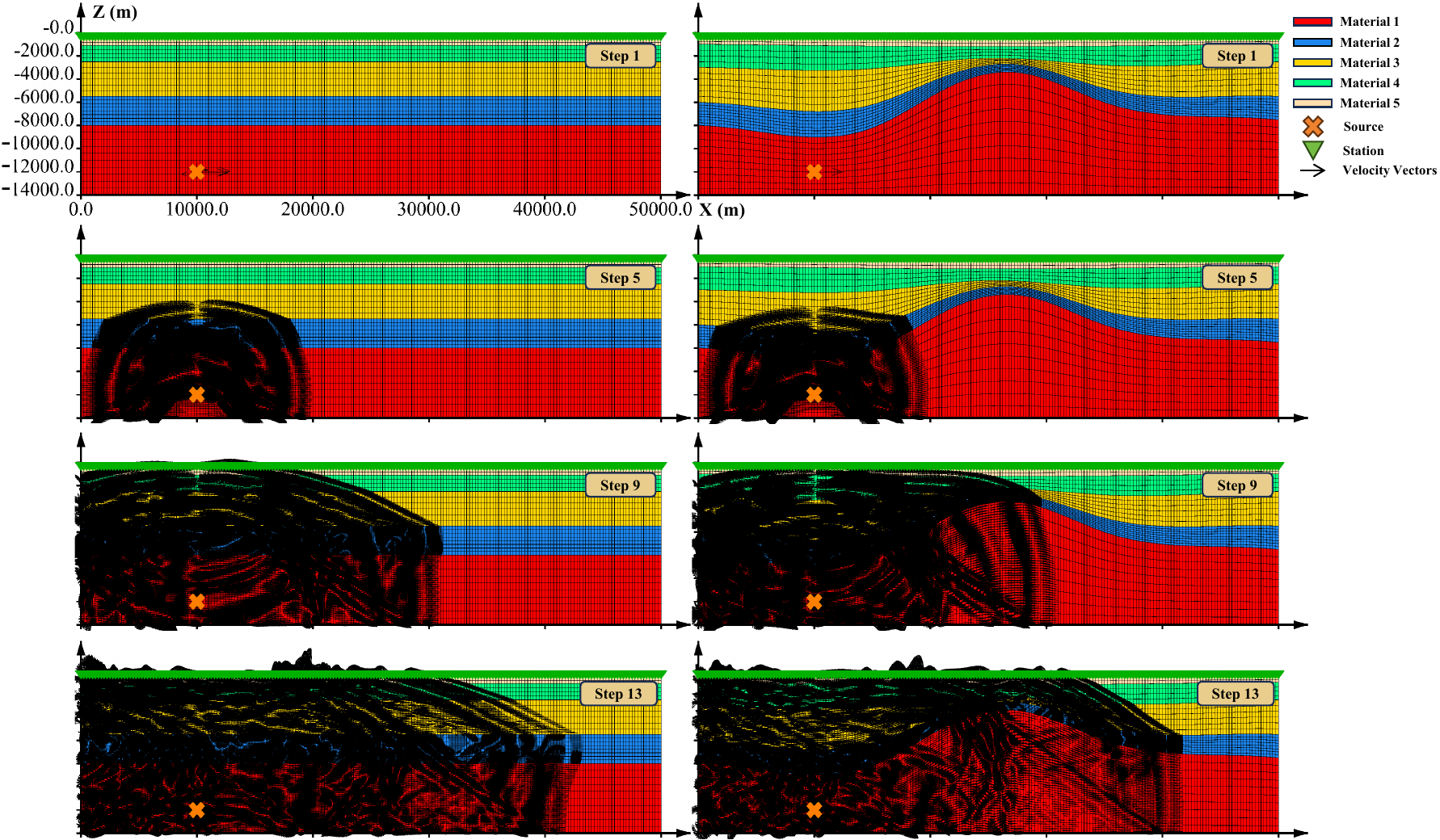}
    \caption{Wavefield snapshots for \textbf{Case~1}, comparing the two models at four time steps. 
    The panels in the \textbf{left column} show Model~A (Flat-Layer Control), while the panels in the \textbf{right column} show Model~B (Buried-Ridge Model).
    Model~B clearly exhibits wavefront acceleration and focusing above the ridge.}
    \label{fig:case1_snapshots}
\end{figure}

\textbf{Wavefield distortion and focusing.}
Figure~\ref{fig:case1_snapshots} compares snapshots of wave propagation for the two models. In 
Model~A, the wavefront spreads normally through the horizontally layered structure, with only minor 
distortion at interface boundaries. In Model~B, the high-velocity basement ridge accelerates the 
central portion of the wavefield, producing a forward-bulging wavefront as it emerges above the ridge. 
This accelerated central arrival is characteristic of seismic focusing through high-velocity 
topography.

\begin{figure}[htbp]
    \centering
    \includegraphics[width=\columnwidth]{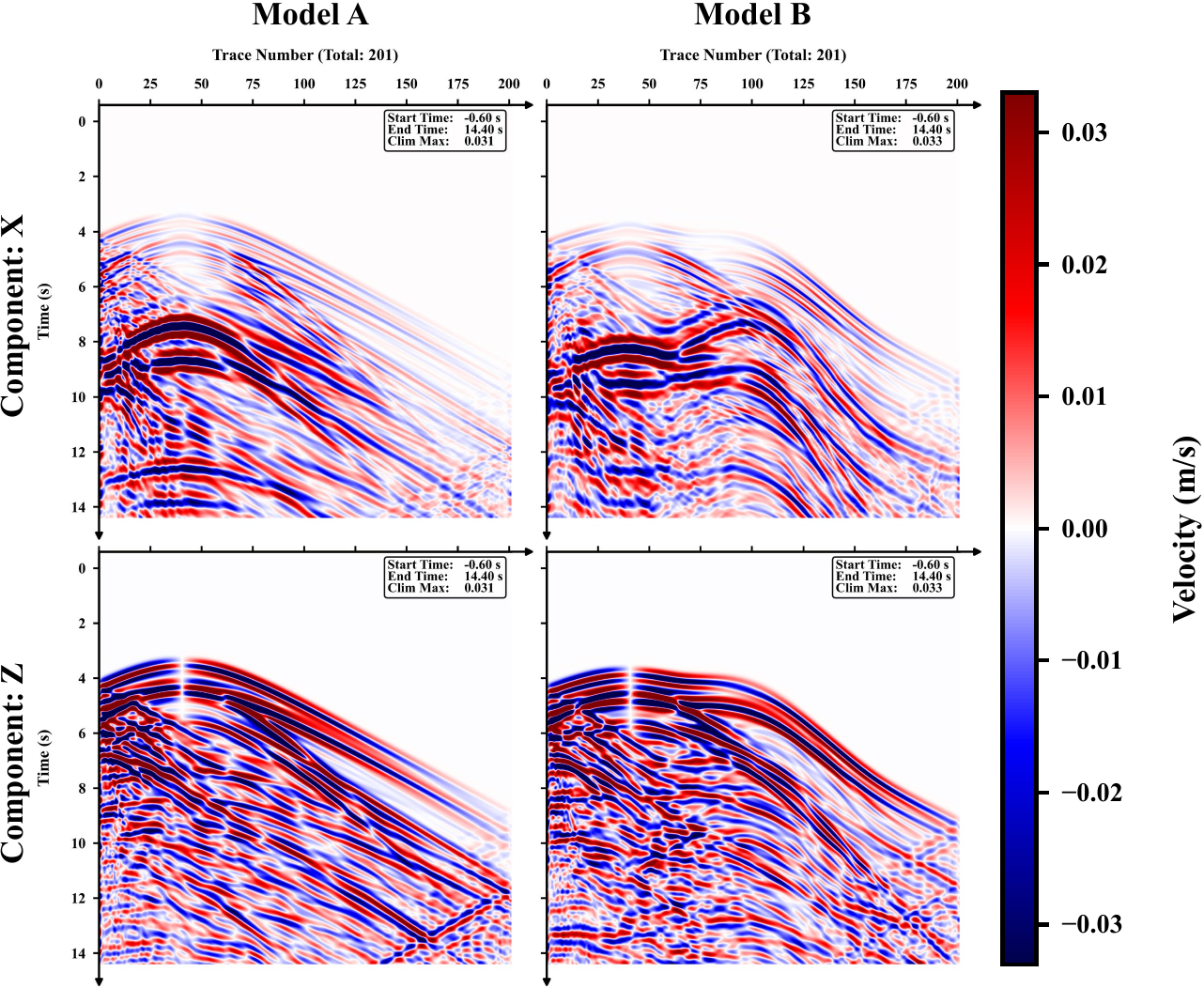}
    \caption{Comparison of surface velocity seismograms (X and Z components) for Model~A and Model~B in \textbf{Case~1}.
    Note that Model~B exhibits earlier arrivals above the buried ridge compared to the flat-layer case (Model~A).}
    \label{fig:case1_seismograms}
\end{figure}

\textbf{Surface expression of the focusing effect.}
The surface seismograms in Figure~\ref{fig:case1_seismograms} show the corresponding differences in 
arrival times. Model~A exhibits nearly linear arrivals across the receiver array, consistent with 
propagation in a flat-layered medium. In Model~B, arrivals in the central portion of the array 
(approximately 20–35~km) occur noticeably earlier, reflecting the faster wave propagation through 
the buried ridge and the resulting wavefront acceleration.

In summary, this case demonstrates that, starting from a single high-level instruction, the agent 
constructs two geometrically distinct models, executes the complete SPECFEM2D workflow, and generates clear visual and seismogram-based evidence of the seismic lens effect, producing a pedagogical comparative experiment.

\subsection{Case 2: High-Level Intent to a Complex 2D Exploration Model—Agent-Led Construction with Light User Guidance}

\subsubsection{Objectives and Setup}
The objective of this case study is to extend the pedagogical experiment of Case~1 to a geologically complex exploration-style scenario. The goal is to evaluate the agent’s capacity to interpret high-level scientific intent and construct a sophisticated 2D P--SV model characterized by realistic stratigraphy and attenuation. To represent a realistic subsurface environment, we construct a $12$~km wide structural model defined by two primary geological components (Figure~\ref{fig:case2_snapshots}).

\begin{itemize}
    \item \textbf{Structural Framework:} A sedimentary sequence comprising nine distinct stratigraphic units, numbered from the surface (Layer~1) down to the basement (Layer~9). The model ($z = -6$ to $0$~km) is characterized by \textbf{strongly undulating, wave-like fold structures} rather than simple flat layers. The surface topography is rugged, featuring a central valley at $x=6$~km flanked by elevated peaks. Material properties exhibit realistic geological variations, with elastic parameters and attenuation factors increasing with depth.

    \item \textbf{Stratigraphic Targets and Anomalies:} The fifth layer exhibits a \textbf{step-like lateral transition}, where the layer interface descends in discrete structural steps towards the right model boundary. A localized reservoir anomaly is embedded within this layer, characterized by low velocity ($V_p=2200$~m/s) and high attenuation ($Q=10$) to simulate a highly absorptive target.
\end{itemize}
The experiment employs a 15~Hz Ricker wavelet source positioned at the surface valley minimum $(x = 6000~\text{m},\, z = -50~\text{m})$. The wavefield response is recorded by a mixed acquisition system comprising a dense surface array (Network AA) of 304 receivers spaced at $\sim40$~m intervals, and two deviated VSP arrays (131 receivers each); the left well (Network BB, $x=4000$ to $2500$~m) deviates away from the center, while the right well (Network CC, $x=8000$ to $6117$~m) converges towards the central anomaly.

\subsubsection{Agent-Driven Interactive Workflow}
The user provides a high-level but descriptive scientific intent:
``Construct a complex 2D P--SV exploration-style forward experiment in SPECFEM2D, completing the full workflow from model building and meshing to forward simulation and visualization. Use a $12$~km wide and $6$~km deep model containing nine strongly undulating, wave-like stratigraphic layers and a rugged surface topography with two peaks and one central valley. Assign geologically reasonable elastic properties and attenuation ($Q$) that increase with depth. Embed a localized reservoir anomaly in Layer~5 with $V_p=2200$~m/s and strong attenuation ($Q=10$), and implement a step-like lateral transition within Layer~5 where the interface drops in discrete steps toward the right. Use a 15~Hz Ricker wavelet source placed near the valley bottom. Deploy a dense surface line of 304 receivers and two deviated VSP arrays (131 receivers each): the left well deviates outward away from the model center, and the right well deviates inward toward the center, but neither well is allowed to intersect the anomaly. Run the simulation on eight GPUs with Stacey absorbing boundaries and wavefield snapshots enabled.''

Upon receiving this instruction, the agent initiates a structured ``Think--Plan--Execute'' procedure through the SPECFEM2D MCP server. After the initial planning phase, the user performs interactive adjustments—specifically, modifying the geometry of the stratigraphic interfaces and refining the distribution of VSP stations.

Subsequently, the agent automatically generates the required SPECFEM2D configuration files, constructs a nine-layer model with strongly undulating interfaces and a dual-peak/central-valley surface, assigns depth-increasing elastic and attenuation parameters, and implements the Layer~5 reservoir anomaly with strong absorption and the prescribed step-like lateral geometry. It then sequentially executes the meshing and forward solvers, configuring parallel computation across eight GPUs with Stacey absorbing boundaries and snapshot output enabled, and organizes the resulting seismograms and wavefield snapshots for visualization and interpretation.

This case highlights the agent’s ability to manage the interdependencies among complex topography, 
layer geometry, multiple material regions, and a multi-component acquisition system—tasks that would 
be tedious, error-prone, and time-consuming to configure manually. It illustrates a clear transition 
from manual operation to intent-driven scientific modeling.

\subsubsection{Results and Discussion}

\begin{figure}[t!]
    \centering
    \includegraphics[width=\columnwidth]{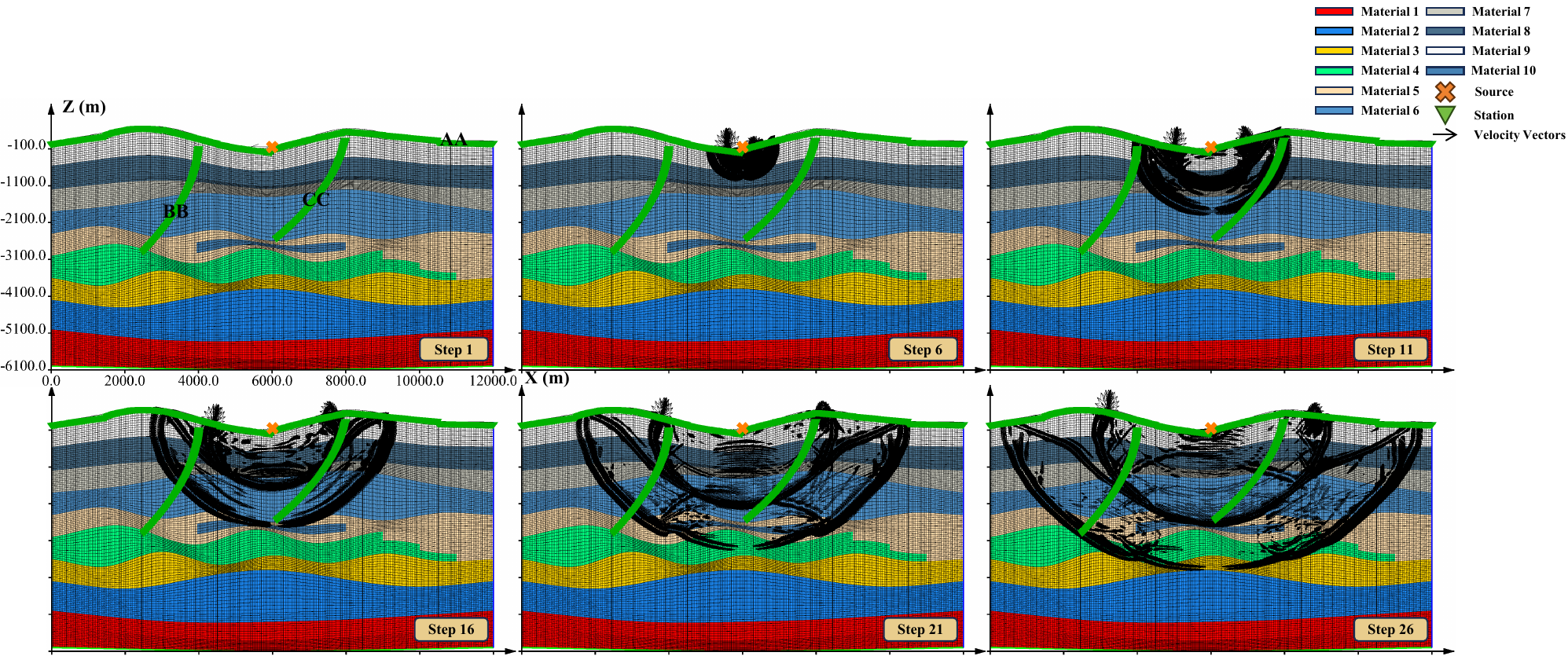}
    \caption{Wavefield snapshots for the complex exploration scenario (\textbf{Case~2}), illustrating the time evolution of the wavefield propagation.
    The labels \textbf{AA}, \textbf{BB}, and \textbf{CC} denote the three receiver arrays.}
    \label{fig:case2_snapshots}
\end{figure}

\begin{figure}[t!]
    \centering
    \includegraphics[width=\columnwidth]{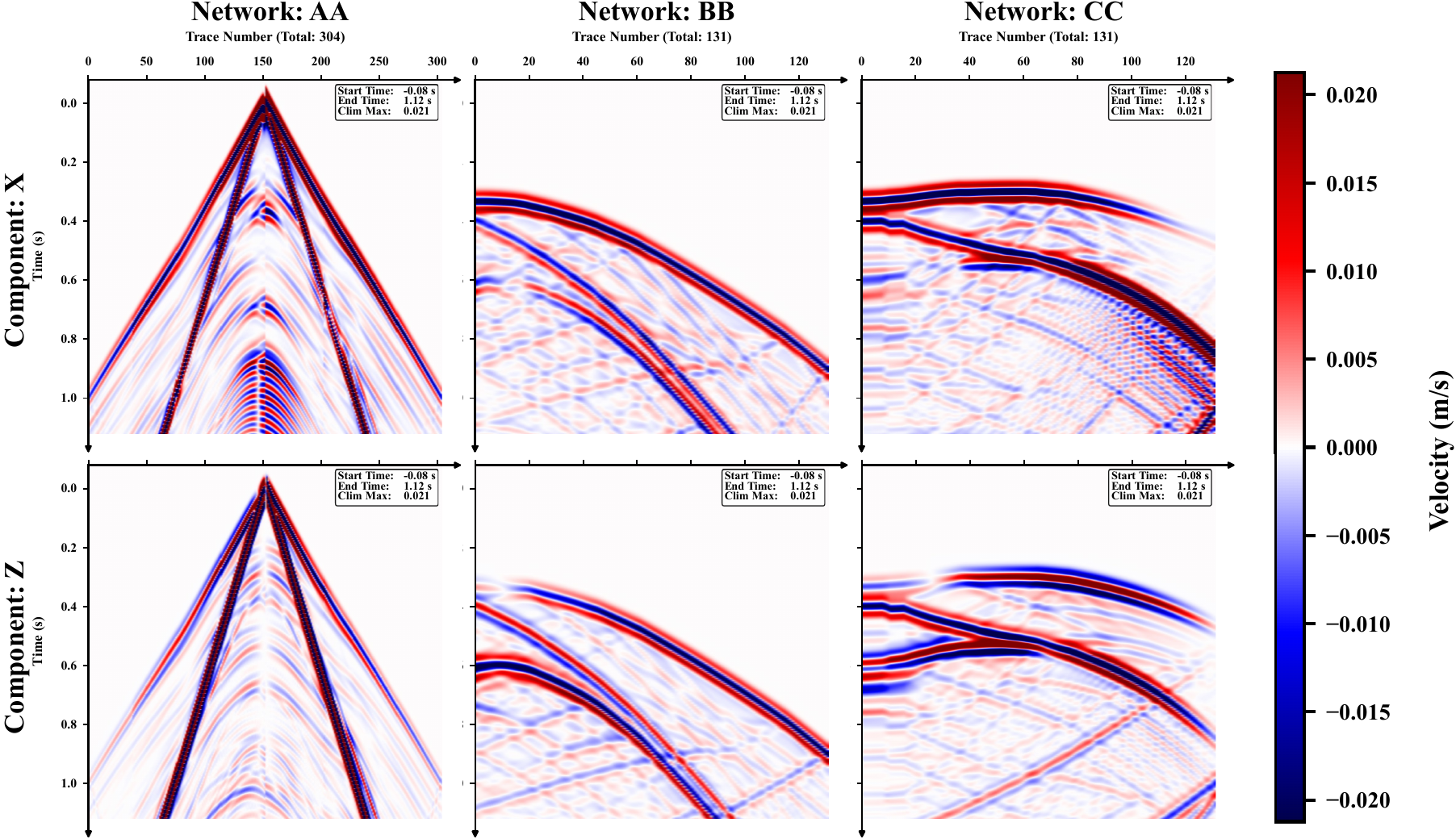}
    \caption{Multi-component velocity seismograms (X and Z) for \textbf{Case~2}, recorded by the surface array (\textbf{AA}), the VSP array extending away from the reservoir anomaly (\textbf{BB}), and the VSP array approaching the anomaly (\textbf{CC}).}
    \label{fig:case2_seismograms}
\end{figure}

The simulation produces wavefield snapshots and multicomponent (X- and Z-component) velocity
records. In this study, we show wavefield snapshots together with the velocity
seismograms, as illustrated in Figures~\ref{fig:case2_snapshots} and \ref{fig:case2_seismograms}.

\textit{Wavefield snapshots.}
Figure~\ref{fig:case2_snapshots} displays several representative snapshots of the simulated
wavefield. The images show the propagation of the wavefront through the undulating stratigraphy
and across the dual-peak surface topography at several time steps.

\textit{Surface and VSP seismograms.}
The velocity seismograms in Figure~\ref{fig:case2_seismograms} show distinct responses
for the surface array and the two VSP wells. The surface receivers capture the general lateral
distribution of the wavefield, whereas the two VSP arrays record depth-dependent responses that
differ from each other due to their relative locations—one well progressively approaches the
reservoir anomaly, while the other moves away from it.

Overall, the results demonstrate that the agent-constructed model and acquisition geometry produce
observable differences in the velocity responses across the surface and borehole
arrays, consistent with the structural variations specified in the user’s high-level request.

\subsection{Case 3: Integrating User-Provided Meshes—Agent-Assisted Completion of the Simulation Workflow}
\subsubsection{Objectives and Setup}
This case evaluates the system’s flexibility in scenarios where part of the modeling workflow has already been completed by the user. Instead of constructing a geological model and mesh from scratch, the objective is to test if the agent can interpret a set of **user-provided external mesh files**, understand their structure, and automatically complete the remaining steps of the forward simulation. To demonstrate this capability in a complex geophysical setting, we utilize a Salt Dome model defined by two primary components (Figure~\ref{fig:case3_snapshots}):

\begin{itemize}
    \item \textbf{External Mesh Parsing:} The input consists of a pre-computed unstructured mesh dataset. The agent is required to ingest these files directly, parsing the topology to identify coordinates, connectivity, and discrete **material indices** (IDs). No physical properties are contained within these files; only the geometric segmentation of regions is provided.

    \item \textbf{Physical Properties and Boundaries:} The simulation configuration maps specific physical parameters to the extracted mesh indices. The model features a high-velocity elastic salt dome (Material~8, $V_p=5000$~m/s) overlain by an acoustic fluid layer (Material~5, $V_s=0$). This setup explicitly addresses the \textbf{fluid-solid coupling} at the interface and incorporates CPML absorbing conditions on the bottom and lateral boundaries to suppress artificial reflections.
\end{itemize}

The experiment employs a 9~Hz Ricker wavelet force source (inclined at $45^\circ$) located deep within the medium at $(x = 2500~\text{m},\, z = 4500~\text{m})$. The wavefield response is recorded by three distinct receiver arrays designed to monitor wave propagation across different media: the sea surface (Network AA), the fluid-solid seabed interface (Network BB), and the deep elastic subsurface (Network CC).

\subsubsection{Agent-Driven Interactive Workflow}
The user provides a high-level scientific intent involving external resources:
``Conduct a forward simulation for a salt dome model with fluid-solid coupling and CPML absorbing boundaries using external CUBIT mesh files provided in a specified directory. Analyze these files to understand the geometric configuration. Assign eight distinct material models, including an acoustic fluid layer ($V_s=0$, Layer~5) and a high-velocity salt dome ($V_p=5000$~m/s, Layer~8). Place the source within the fluid layer. Deploy three receiver arrays: one at the sea surface, one at the seafloor, and one deep in the subsurface beneath the salt dome, ensuring the deep array does not intersect the salt body.''

Upon receiving this instruction, the agent initiates the ``Think--Plan--Execute'' procedure through the SPECFEM2D MCP server. It first utilizes file system tools to inspect and analyze the provided mesh files, autonomously inferring the model domain and geometric constraints. Following this analysis, the user performs interactive adjustments, focusing primarily on refining the spatial distribution of the three receiver arrays to ensure they align with the fluid interfaces and avoid the salt body as requested.

Subsequently, the agent automatically generates the compatible SPECFEM2D configuration files, strictly adhering to the fixed topology of the external mesh and the specified material properties. It configures the simulation for fluid-solid coupling with CPML boundaries, executes the solver, and organizes the resulting wavefield snapshots and seismograms for visualization.

This case demonstrates the agent's ability to integrate seamlessly into workflows that start from user-generated meshes and to avoid redundant model-building stages.

\subsubsection{Results and Discussion}

\begin{figure}[t!]
    \centering
    \includegraphics[width=\columnwidth]{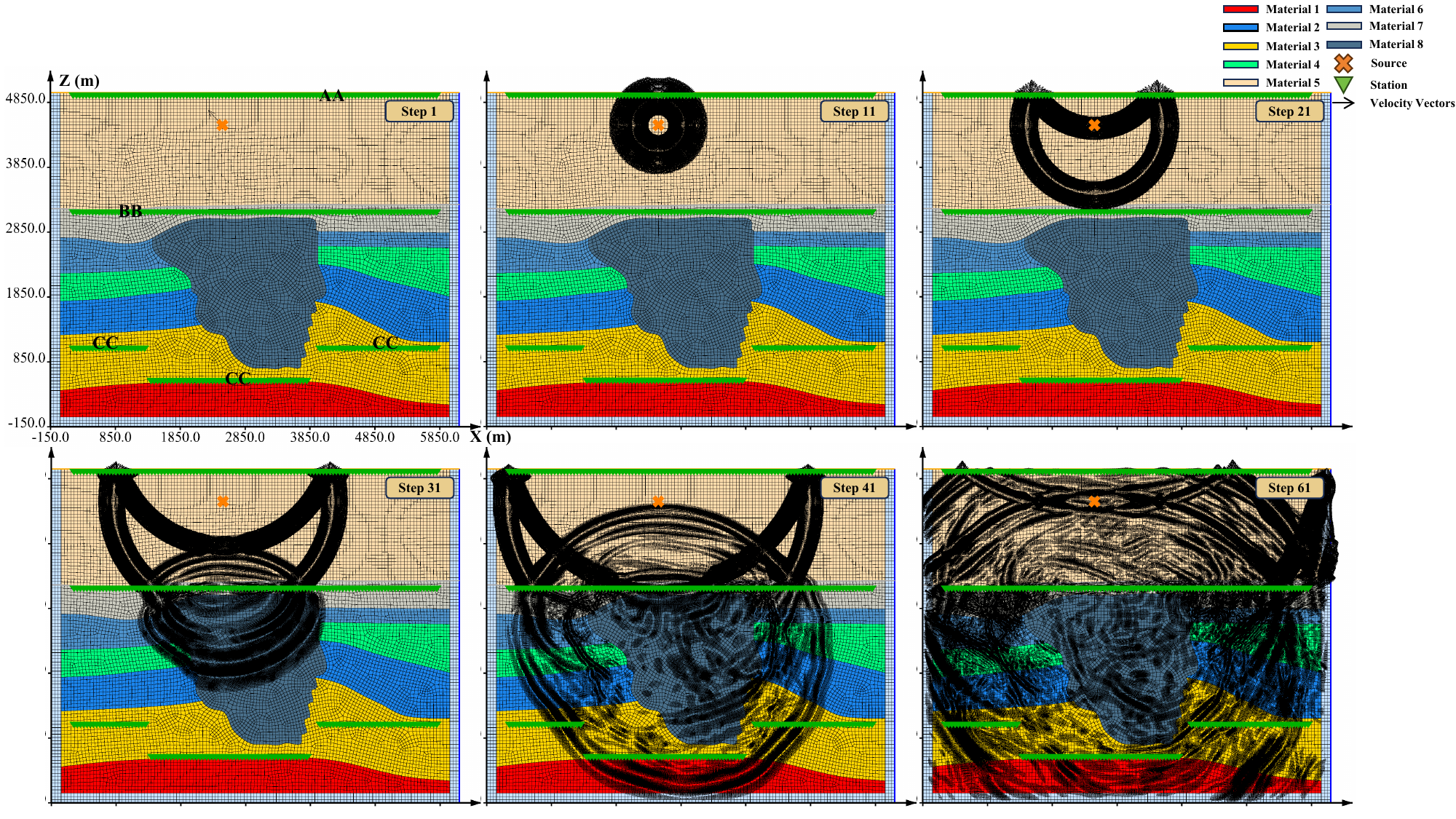}
    \caption{Wavefield snapshots for the Salt Dome benchmark model (\textbf{Case~3}).
    This simulation utilizes a \textbf{user-provided mesh} to accurately handle the \textbf{fluid-solid coupling} between the top \textbf{acoustic water layer} and the elastic subsurface. 
    The receiver networks correspond to the sea surface (\textbf{AA}), the seabed (\textbf{BB}), and the subsurface (\textbf{CC}), respectively.}
    \label{fig:case3_snapshots}
\end{figure}

The simulation completes successfully using the user-provided mesh. Figure~\ref{fig:case3_snapshots} shows representative wavefield snapshots at several time steps, illustrating the wavefield propagation through the acoustic fluid layer, the transmission across the fluid-solid seafloor interface, and the complex interaction with the high-velocity salt dome.

\begin{figure}[t!]
    \centering
    \includegraphics[width=\columnwidth]{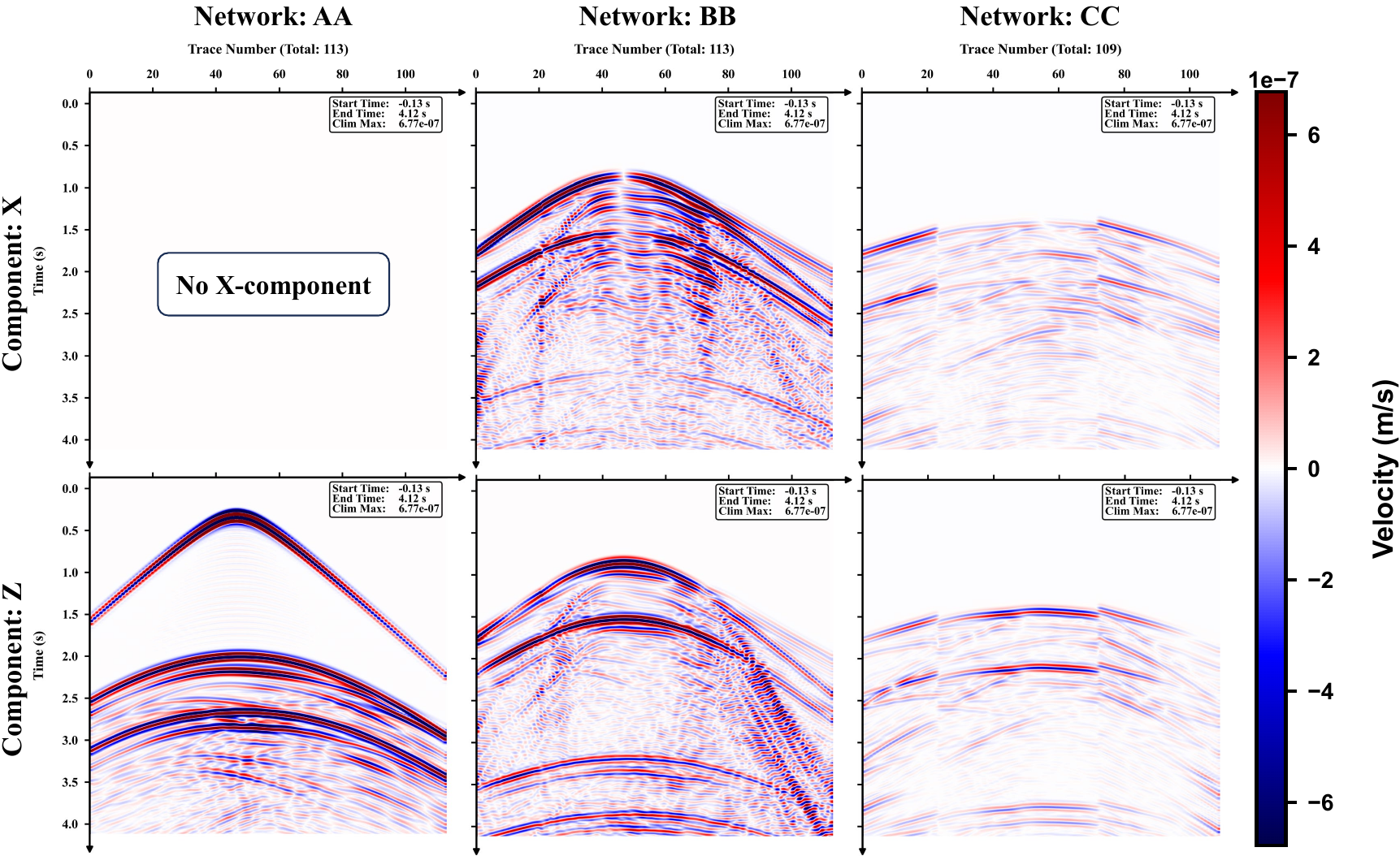}
    \caption{Multi-component velocity seismograms (X and Z) for the Salt Dome benchmark model (\textbf{Case~3}), recorded by the receiver networks at the sea surface (\textbf{AA}), on the seabed (\textbf{BB}), and within the subsurface (\textbf{CC}). 
    Notably, the horizontal (X) component at the sea surface (\textbf{AA}) is zero, consistent with the free-surface boundary condition for an acoustic fluid.}
    \label{fig:case3_seismograms}
\end{figure}

The velocity seismograms in Figure~\ref{fig:case3_seismograms} present recordings from three distinct receiver arrays: (a) at the sea surface (\textbf{AA}), (b) on the seabed (\textbf{BB}), and (c) within the deep subsurface (\textbf{CC}). These datasets display distinct response patterns due to their specific locations relative to the fluid interfaces and the salt body. Specifically, the receiver array at the sea surface (\textbf{AA}) exhibits a vanishing horizontal (X) component. This observation aligns perfectly with the physical characteristics of an acoustic fluid layer, which does not support shear wave propagation, thereby validating the correct implementation of the acoustic-elastic coupling. In contrast, the seabed (\textbf{BB}) and subsurface (\textbf{CC}) arrays record significant energy on both components, capturing the complex mode conversions at the seabed interface and the scattering effects induced by the salt dome.

Overall, this case demonstrates that the agent can successfully complete the SPECFEM2D workflow using user-provided mesh files, automatically generating missing configuration files and executing the simulation and visualization process with minimal user intervention.

\subsection{Case 4: 3D Forward Simulation of the Campi Flegrei Volcanic Region—Agent-Assisted Configuration and Execution}

\subsubsection{Objectives and Setup}
\begin{figure}[htbp]
    \centering
    \includegraphics[width=\columnwidth]{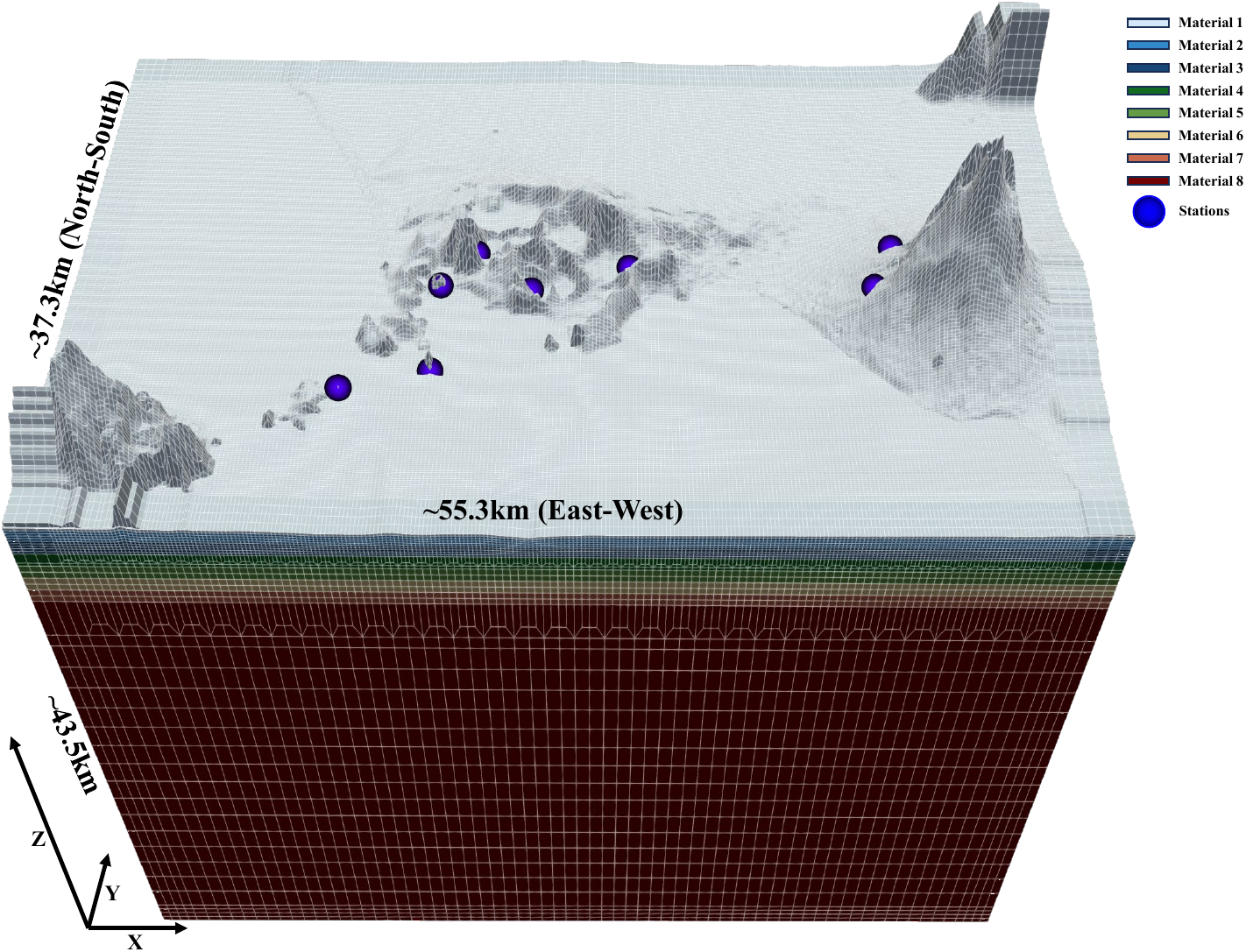}
    \caption{External view of the 3D computational mesh for the Campi Flegrei region (\textbf{Case~4}). 
    \textbf{Blue spheres} indicate the locations of the \textbf{nine} surface stations (note that one station is occluded in this perspective). 
    The topography is \textbf{vertically exaggerated} for better visualization of the surface relief.}
    \label{fig:case4_mesh}
\end{figure}

This case evaluates the agent’s ability to configure and execute a 3D forward simulation for a real-world volcanic environment using SPECFEM3D\_Cartesian. The simulation targets the Campi Flegrei region in Italy, reproducing a specific experimental configuration \cite{SPECFEM3D_GitHubDevel}. To represent the complex volcanic setting accurately, we construct a high-fidelity regional model defined by two primary components (Figure~\ref{fig:case4_mesh}):

\begin{itemize}
    \item \textbf{Computational Domain and Physics:} The simulation targets the Campi Flegrei volcanic region in Italy, configured within the UTM Zone 33N projection system. The model incorporates realistic surface topography to accurately capture topographic scattering effects. To ensure physical realism, we employ a 1D layered velocity model combined with viscoelastic attenuation (Olsen's model), utilizing Convolutional Perfectly Matched Layers on the lateral and bottom boundaries to absorb outgoing energy.

    \item \textbf{Source and Observation Configuration:} The experiment models a shallow volcanic seismic event located at a depth of 2.6~km ($40.8277^\circ$N, $14.1380^\circ$E). The source is defined by a full 3D Moment Tensor (CMT) solution with a half-duration of 0.1025~s. The resulting wavefield is recorded by a realistic monitoring network consisting of nine INGV surface stations, which are irregularly distributed across the caldera to capture the velocity response of the complex medium.
\end{itemize}

Key elements of the user-specified configuration include:  
(1) a hexahedral spectral-element mesh, a timestep of 0.001~s, and a total duration of 40~s;  
(2) attenuation enabled using an Olsen-type model;  
(3) C-PML absorbing boundaries except for a free surface;  
(4) generation of three configuration files—\texttt{Par\_file}, \texttt{CMTSOLUTION}, and 
\texttt{STATIONS};  
(5) a shallow volcanic source at 2.6~km depth with a specified moment tensor; and  
(6) nine surface stations from the INGV monitoring network.

The computational mesh used for the simulation is shown in 
Figure~\ref{fig:case4_mesh}.

\subsubsection{Agent-Driven Interactive Workflow}
The user provides a comprehensive technical specification based on a real-world experimental configuration:
Configure and execute a high-fidelity 3D forward simulation for the Campi Flegrei volcanic region. The user provides existing interface files, which the agent is required to inspect and analyze automatically. Furthermore, the specification details the regional settings and projection, time step parameters, mesh configuration, material properties, boundary conditions, attenuation settings, source parameters, station configuration, and visualization parameters.

Upon receiving this detailed instruction, the agent initiates the ``Think--Plan--Execute'' procedure through the SPECFEM3D\_Cartesian MCP server. Unlike scenarios requiring autonomous inference, the agent's primary role here is to strictly interpret and parse the user's rigorous parameter set. Following the initial parsing, the workflow proceeds through multi-turn interactions. Subsequently, the agent generates the configuration files and automatically executes the forward simulation workflow. Finally, it reads the output directory and automatically completes the visualization of the velocity seismograms.

This case demonstrates the agent's capability, combined with the MCP server, to handle experiments in real-world 3D scenarios. It converts a dense, user-provided experimental design document into an executable simulation environment while allowing for human supervision to ensure precision in critical engineering parameters.

\subsubsection{Results and Discussion}

\begin{figure}[htbp]
    \centering
    \includegraphics[width=\columnwidth]{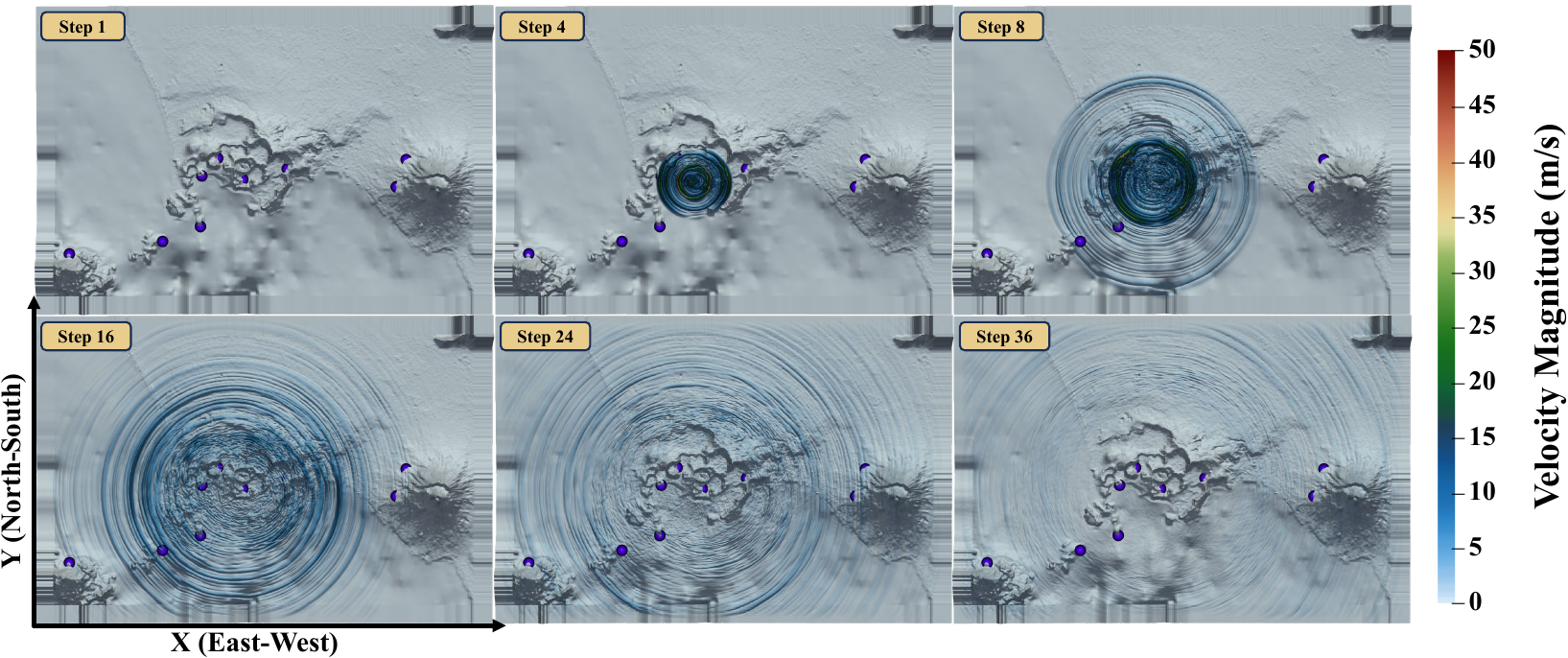}
    \caption{Surface wavefield snapshots for the Campi Flegrei model (\textbf{Case~4}), illustrating the time evolution of wave propagation across the complex topography.}
    \label{fig:case4_snapshots}
\end{figure}

The simulation completes successfully using the generated configuration. 
Figure~\ref{fig:case4_snapshots} shows representative snapshots of the surface wavefield at several 
time steps, illustrating the propagation of the simulated wavefield across the model surface.

\begin{figure}[htbp]
    \centering
    \includegraphics[width=\columnwidth]{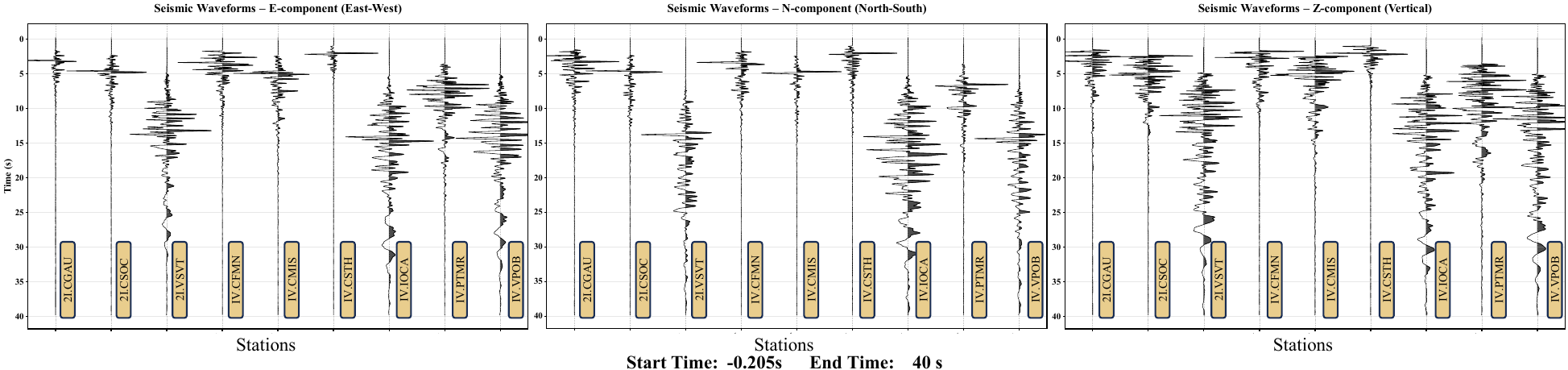}
    \caption{Three-component velocity seismograms (East, North, and Vertical) for the Campi Flegrei model (\textbf{Case~4}), recorded by the \textbf{nine} surface stations.\textbf{Positive amplitudes are filled in black}, while negative amplitudes remain white.}
    \label{fig:case4_seismograms}
\end{figure}

The velocity seismograms recorded at the nine stations 
(Figure~\ref{fig:case4_seismograms}) display distinct responses across the array due to the differing 
station positions relative to the shallow volcanic source.

Overall, this case demonstrates that the agent can correctly interpret a detailed 3D modeling 
instruction, generate all necessary SPECFEM3D\_Cartesian configuration files, run a multi-process 
simulation, and produce the required waveform and visualization outputs with user 
intervention.

\subsection{Case 5: Global Forward Simulation with SPECFEM3D\_GLOBE—Agent-Assisted Configuration and Execution}

\subsubsection{Objectives and Setup}
\begin{figure}[htbp]
    \centering
    \includegraphics[width=\columnwidth]{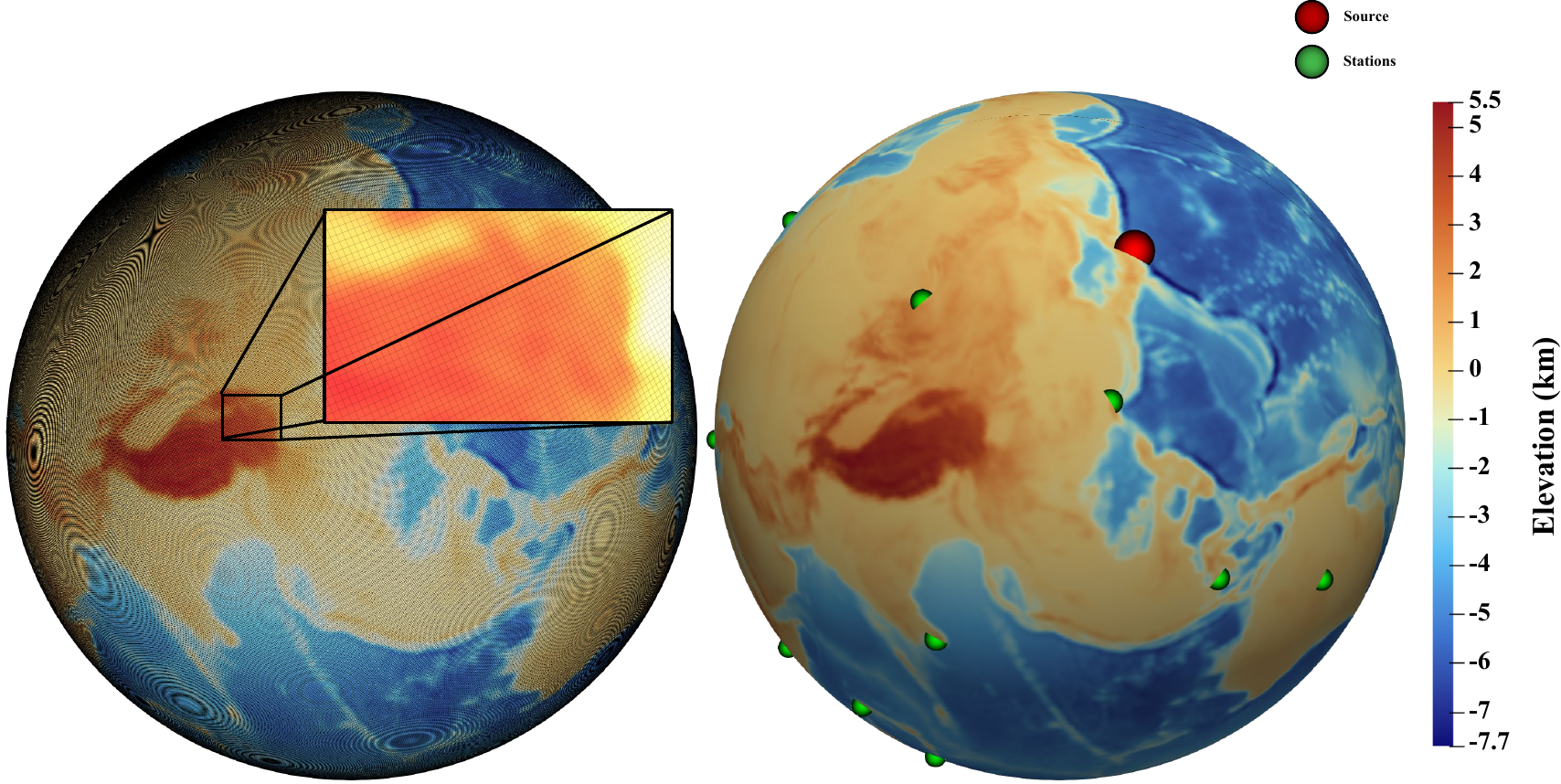}
    \caption{Global mesh configuration and observation system for the s362ani Earth model (\textbf{Case~5}), simulating an $M_w$ 9.1 earthquake scenario in the Tohoku region. 
    \textbf{Left}: Visualization of the global cubed-sphere spectral-element mesh, including a \textbf{zoom-in inset} to illustrate the high-resolution grid discretization.
    \textbf{Right}: Spatial distribution of the twelve global seismic stations (green spheres) and the seismic source (red sphere). Note that stations located on the \textbf{far side of the globe} are not visible in this view.}
    \label{fig:case5_mesh}
\end{figure}

The objective of this case study is to evaluate the framework’s capability to handle a realistic, planetary-scale forward simulation using SPECFEM3D\_GLOBE. This case involves high-fidelity global modeling with full physical complexity. To simulate a realistic scenario, we configure a global Earth model based on the s362ani 3D mantle structure \cite{Kustowski2008s362ani} (Figure~\ref{fig:case5_mesh}).

\begin{itemize}
    \item \textbf{Global Mesh and Physics:} The computational domain is discretized using a global spectral-element mesh consisting of six chunks with a resolution of $NEX=64$ elements per chunk. To ensure realistic wave propagation, we enable full physical complexities, including self-gravity, Earth's rotation, attenuation, crustal topography, and explicit ocean loads.
    \item \textbf{Source and Observation System:} The simulation models a Tohoku region earthquake ($M_w$ 9.1). The source is defined by a Centroid Moment Tensor (CMT) solution placed at a depth of 20~km ($38.30^\circ$N, $142.37^\circ$E), with a half-duration of 15.0~s. The resulting wavefield is recorded by a global network of 12 distinct stations from the II and IU networks, distributed at varying elevations and burial depths.
\end{itemize}

\subsubsection{Agent-Driven Interactive Workflow}

The user provides a comprehensive high-level instruction specifying the hardware resources (a single NVIDIA A800 GPU), the scientific scenario (the 2011 Tohoku earthquake), and the context of the local environment (availability of s362ani, crust2.0, topo\_bathy, and QRFSI12 data). Upon receiving this request through the SPECFEM3D\_GLOBE MCP server, the agent initiates a ``Think--Plan--Execute'' cycle. It autonomously performs a web search to retrieve the necessary parameters for the 2011 Tohoku-Oki earthquake, obtaining the precise hypocenter coordinates, moment tensor components, and information regarding the distribution of representative global stations.

Following a multi-turn dialogue with the user to refine and confirm these configurations, the agent automatically generates the required input files: a \texttt{Par\_file} configured for GPU execution (\texttt{GPU\_MODE = .true.}) with full physical complexities enabled, a \texttt{CMTSOLUTION} file populated with the historical event data, and a \texttt{STATIONS} file. Subsequently, it launches the meshing and forward simulation executables on the specified GPU hardware. Finally, upon completion of the simulation, the agent autonomously scans the output directory, reads the generated station signal files, and produces the visualization plots of the station signals.

\subsubsection{Results and Discussion}

\begin{figure}[htbp]
    \centering
    \includegraphics[width=\columnwidth]{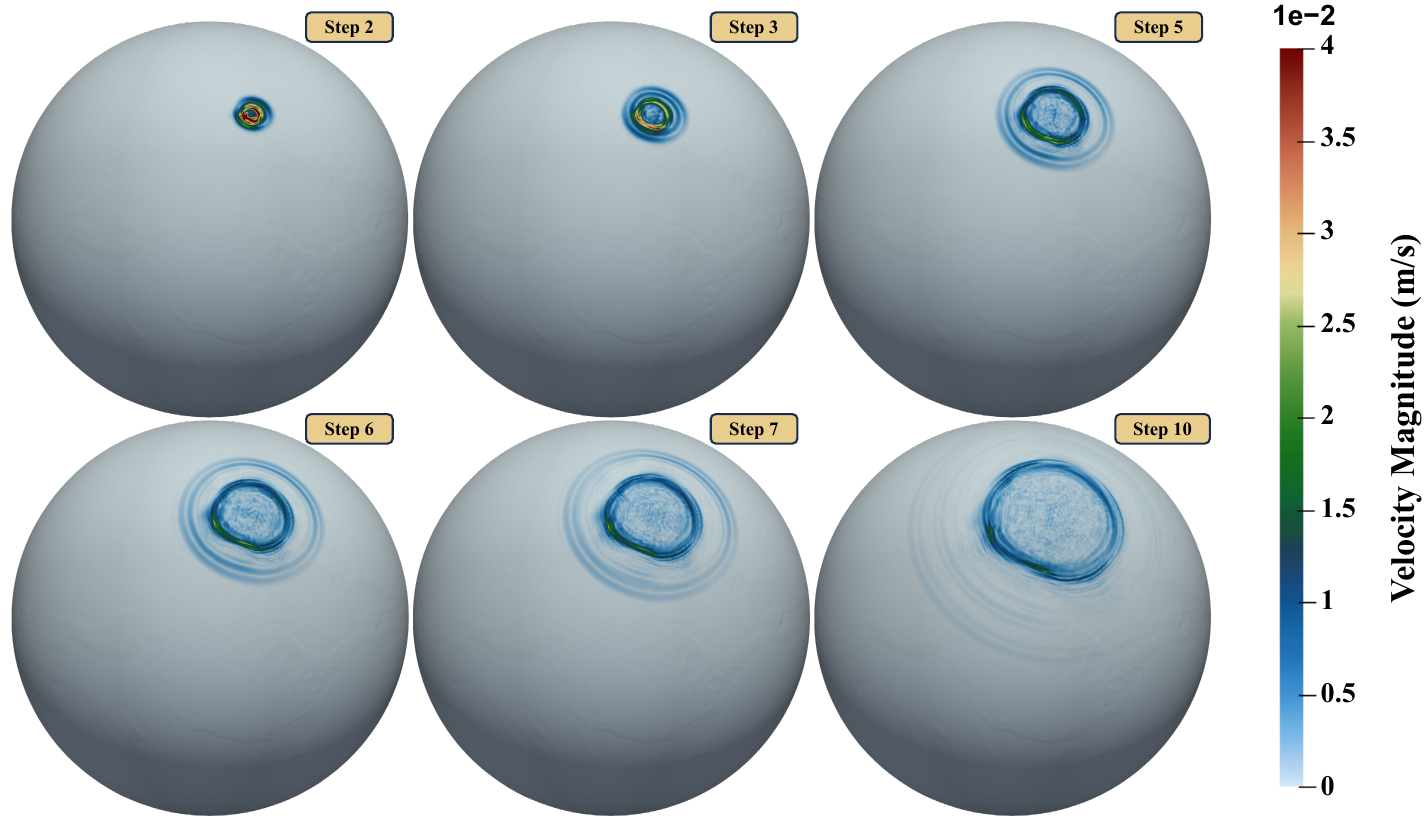}
    \caption{Global surface wavefield snapshots for the s362ani model (\textbf{Case~5}), illustrating the time evolution of seismic wave propagation across the globe.}
    \label{fig:case5_snapshots}
\end{figure}

The simulation completes successfully, leveraging the GPU architecture to propagate seismic waves through the complex global model.
Figure~\ref{fig:case5_snapshots} presents representative global wavefield snapshots at several time 
steps, illustrating large-scale wave propagation around the planet.

\begin{figure}[htbp]
    \centering
    \includegraphics[width=\columnwidth]{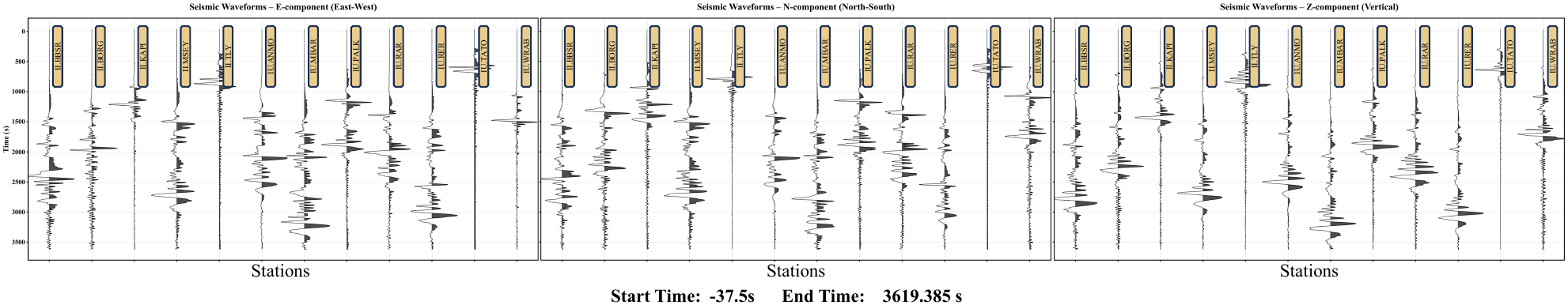}
    \caption{Three-component velocity seismograms (North, East, and Vertical) for the global earthquake scenario (\textbf{Case~5}), recorded by the \textbf{twelve} global stations.\textbf{Positive amplitudes are filled in black}, while negative amplitudes remain white.}
    \label{fig:case5_seismograms}
\end{figure}

As shown in Figure~\ref{fig:case5_seismograms}, the velocity seismograms at the
twelve globally distributed stations exhibit strong motion almost immediately
after the simulation start. The figure presents the
waveforms for all twenty-four events.
The velocity seismograms in Figure~\ref{fig:case5_seismograms} show the three-component recordings at the twelve globally distributed stations. Unlike simple synthetic tests, these waveforms exhibit complex phase arrivals and amplitude variations. The successful retrieval of these waveforms validates the agent's ability to correctly configure high-fidelity, GPU-accelerated global simulations from high-level instructions.

\section{Conclusion and Outlook}
\label{sec:Conclusion and Outlook}
In this paper, we introduce an innovative workflow for geophysical forward simulation that deeply
integrates LLMs with traditional computational software. By developing a
dedicated suite of MCP servers for SPECFEM, we effectively leverage LLMs to
assist and automate the complete lifecycle of seismic simulations.

Our five case studies confirm that this agentic workflow represents a powerful paradigm shift in the
use of geophysical software. The LLM-driven system enables researchers to issue commands and
interact using natural language, while the AI handles the complex underlying execution—ranging from
simple 2D scenarios to realistic 3D and global-scale models. This significantly lowers the software
learning curve and reduces manual effort, allowing researchers to work more intuitively and
efficiently by focusing on their scientific objectives rather than technical implementation details.

Beyond mere efficiency, this work establishes a foundation for the next generation of automated
scientific research. We envision a future where intelligent agents evolve from passive assistants
into active research partners capable of autonomous discovery. Realizing this vision requires not
only the integration of domain-specific tools, but also the incorporation of geophysical domain
knowledge and the agent’s ability to make its own decisions, learn strategies, and summarize
experience through iterative interaction with the workflow. In such a setting, agents gain the
potential to manage fully closed-loop research cycles: autonomously generating hypotheses,
constructing subsurface models, designing observation geometries, executing forward simulations and
inversion workflows, analyzing residuals, and performing geological interpretation across the entire
seismic exploration process. Without human intervention, the agent can iteratively optimize
acquisition systems, numerical algorithms, hyperparameters, and subsurface models. This shift toward
autonomous geophysical laboratories promises to accelerate the exploration of vast parameter spaces
and reveal complex geological patterns that remain difficult to uncover using traditional manual
workflows.

The agent-driven research paradigm marks the arrival of an era where scientists and intelligent
agents collaborate to solve complex problems. Future research will focus on expanding the MCP server
functionalities to integrate a broader ecosystem of heterogeneous geophysical software—with diverse
formats and capabilities, including but not limited to preprocessing, postprocessing, imaging,
inversion, and interpretation tools—achieving a deep integration of the agent with seismic inversion
workflows, and enhancing the agent’s abilities for error correction, exploratory learning, and
experience summarization so that it can robustly handle unexpected simulation challenges, ultimately
moving toward the vision of fully autonomous geophysical research.

\bibliography{ref}

@book{AkiRichards2002,
  title={Quantitative seismology},
  author={Aki, Keiiti and Richards, Paul G},
  year={2002}
}

@article{Patera1984,
  title={A spectral element method for fluid dynamics: laminar flow in a channel expansion},
  author={Patera, Anthony T},
  journal={Journal of computational Physics},
  volume={54},
  number={3},
  pages={468--488},
  year={1984},
  publisher={Elsevier}
}

@article{SerianiPriolo1994,
  title={Spectral element method for acoustic wave simulation in heterogeneous media},
  author={Seriani, G{\'e}za and Priolo, Enrico},
  journal={Finite elements in analysis and design},
  volume={16},
  number={3-4},
  pages={337--348},
  year={1994},
  publisher={Elsevier}
}

@article{KomatitschVilotte1998,
  title={The spectral element method: an efficient tool to simulate the seismic response of 2D and 3D geological structures},
  author={Komatitsch, Dimitri and Vilotte, Jean-Pierre},
  journal={Bulletin of the seismological society of America},
  volume={88},
  number={2},
  pages={368--392},
  year={1998},
  publisher={The Seismological Society of America}
}

@article{KomatitschTromp1999,
  title={Introduction to the spectral element method for three-dimensional seismic wave propagation},
  author={Komatitsch, Dimitri and Tromp, Jeroen},
  journal={Geophysical journal international},
  volume={139},
  number={3},
  pages={806--822},
  year={1999},
  publisher={Blackwell Publishing Ltd Oxford, UK}
}

@article{KomatitschTromp2002a,
  title={Spectral-element simulations of global seismic wave propagation—I. Validation},
  author={Komatitsch, Dimitri and Tromp, Jeroen},
  journal={Geophysical Journal International},
  volume={149},
  number={2},
  pages={390--412},
  year={2002},
  publisher={Blackwell Publishing Ltd Oxford, UK}
}

@article{KomatitschTromp2002b,
  author  = {Komatitsch, Dimitri and Tromp, Jeroen},
  title   = {Spectral-element simulations of global seismic wave propagation -- II. 3-D models, oceans, rotation and self-gravitation},
  journal = {Geophysical Journal International},
  year    = {2002},
  volume  = {150},
  number  = {1},
  pages   = {303--318}
}

@article{TrompKomatitschLiu2008,
  title={Spectral-element and adjoint methods in seismology},
  author={Tromp, Jeroen and Komatitsch, Dimitri and Liu, Qinya and others},
  journal={Communications in Computational Physics},
  volume={3},
  number={1},
  pages={1--32},
  year={2008}
}

@book{Fichtner2010,
  title={Full seismic waveform modelling and inversion},
  author={Fichtner, Andreas},
  year={2010},
  publisher={Springer Science \& Business Media}
}

@article{KomatitschMartin2007,
  title={An unsplit convolutional perfectly matched layer improved at grazing incidence for the seismic wave equation},
  author={Komatitsch, Dimitri and Martin, Roland},
  journal={Geophysics},
  volume={72},
  number={5},
  pages={SM155--SM167},
  year={2007},
  publisher={Society of Exploration Geophysicists}
}

@misc{Carrington2008,
  title={High-frequency simulations of global seismic wave propagation using SPECFEM3D\_GLOBE on 62 thousand processor cores. In 2008 SC-International Conference for High Performance Computing, Networking, Storage and Analysis (pp. 1--11). Austin, TX: IEEE},
  author={Carrington, L and Komatitsch, D and Laurenzano, M and Tikir, M and Mich{\'e}a, D and Le Goff, N and Tromp, J},
  year={2008}
}

@article{Peter2011,
  title={Forward and adjoint simulations of seismic wave propagation on fully unstructured hexahedral meshes},
  author={Peter, Daniel and Komatitsch, Dimitri and Luo, Yang and Martin, Roland and Le Goff, Nicolas and Casarotti, Emanuele and Le Loher, Pieyre and Magnoni, Federica and Liu, Qinya and Blitz, C{\'e}line and others},
  journal={Geophysical Journal International},
  volume={186},
  number={2},
  pages={721--739},
  year={2011},
  publisher={Blackwell Publishing Ltd Oxford, UK}
}

@article{Kustowski2008s362ani,
  title={Anisotropic shear-wave velocity structure of the Earth's mantle: A global model},
  author={Kustowski, B and Ekstr{\"o}m, G and Dziewo{\'n}ski, AM},
  journal={Journal of Geophysical Research: Solid Earth},
  volume={113},
  number={B6},
  year={2008},
  publisher={Wiley Online Library}
}

@misc{SPECFEMWeb2025,
  author       = {{SPECFEM Developers}},
  title        = {SPECFEM: A family of open-source spectral-element method solvers},
  howpublished = {\url{https://specfem.org/}},
  year         = {2025},
  note         = {Accessed 7 December 2025}
}

@misc{SPECFEM3DCartesianManual,
  author       = {Komatitsch, Dimitri and Tromp, Jeroen and Peter, Daniel and others},
  title        = {SPECFEM3D Cartesian User Manual},
  howpublished = {\url{https://specfem3d.readthedocs.io/}},
  year         = {2010},
  note         = {Online documentation, accessed 7 December 2025}
}

@inproceedings{Yao2023ReAct,
  title={React: Synergizing reasoning and acting in language models},
  author={Yao, Shunyu and Zhao, Jeffrey and Yu, Dian and Du, Nan and Shafran, Izhak and Narasimhan, Karthik R and Cao, Yuan},
  booktitle={The eleventh international conference on learning representations},
  year={2022}
}

@article{Wang2023PlanSolve,
  title={Plan-and-solve prompting: Improving zero-shot chain-of-thought reasoning by large language models},
  author={Wang, Lei and Xu, Wanyu and Lan, Yihuai and Hu, Zhiqiang and Lan, Yunshi and Lee, Roy Ka-Wei and Lim, Ee-Peng},
  journal={arXiv preprint arXiv:2305.04091},
  year={2023}
}

@article{Chen2025ScienceAgentBench,
  title={Scienceagentbench: Toward rigorous assessment of language agents for data-driven scientific discovery},
  author={Chen, Ziru and Chen, Shijie and Ning, Yuting and Zhang, Qianheng and Wang, Boshi and Yu, Botao and Li, Yifei and Liao, Zeyi and Wei, Chen and Lu, Zitong and others},
  journal={arXiv preprint arXiv:2410.05080},
  year={2024}
}

@article{OpenAI2025PaperBench,
  title={PaperBench: Evaluating AI's Ability to Replicate AI Research},
  author={Starace, Giulio and Jaffe, Oliver and Sherburn, Dane and Aung, James and Chan, Jun Shern and Maksin, Leon and Dias, Rachel and Mays, Evan and Kinsella, Benjamin and Thompson, Wyatt and others},
  journal={arXiv preprint arXiv:2504.01848},
  year={2025}
}

@article{Swanson2025VirtualLab,
  title={The Virtual Lab of AI agents designs new SARS-CoV-2 nanobodies},
  author={Swanson, Kyle and Wu, Wesley and Bulaong, Nash L and Pak, John E and Zou, James},
  journal={Nature},
  volume={646},
  number={8085},
  pages={716--723},
  year={2025},
  publisher={Nature Publishing Group UK London}
}

@misc{Anthropic2024MCP,
  author       = {{Anthropic}},
  title        = {Introducing the Model Context Protocol},
  howpublished = {\url{https://www.anthropic.com/news/model-context-protocol}},
  year         = {2024},
  note         = {Accessed 7 December 2025}
}

@misc{MCPSpec2025,
  author       = {{Model Context Protocol}},
  title        = {Model Context Protocol Specification (Protocol Revision 2025-11-25)},
  howpublished = {\url{https://modelcontextprotocol.io/specification/2025-11-25}},
  year         = {2025},
  note         = {Accessed 7 December 2025}
}

@misc{GoogleCloud2025MCP,
  author       = {{Google Cloud}},
  title        = {What is Model Context Protocol (MCP)? A guide},
  howpublished = {\url{https://cloud.google.com/discover/what-is-model-context-protocol}},
  year         = {2025},
  note         = {Accessed 7 December 2025}
}

@misc{ClineWebsite2025,
  author       = {{Cline Team}},
  title        = {Cline: The Open Coding Agent},
  howpublished = {\url{https://cline.bot/}},
  year         = {2025},
  note         = {Accessed 7 December 2025}
}

@misc{ClineVSCode2024,
  author       = {Rizwan, Saoud},
  title        = {Cline -- AI Coding Agent for VS Code},
  howpublished = {\url{https://marketplace.visualstudio.com/items?itemName=saoudrizwan.claude-dev}},
  year         = {2024},
  note         = {VS Code Marketplace entry, accessed 7 December 2025}
}

@misc{MCPClients2025,
  author       = {{Model Context Protocol}},
  title        = {Example Clients: Cline},
  howpublished = {\url{https://modelcontextprotocol.io/clients}},
  year         = {2025},
  note         = {Accessed 7 December 2025}
}

@article{durante2024agent,
  title={Agent ai: Surveying the horizons of multimodal interaction},
  author={Durante, Zane and Huang, Qiuyuan and Wake, Naoki and Gong, Ran and Park, Jae Sung and Sarkar, Bidipta and Taori, Rohan and Noda, Yusuke and Terzopoulos, Demetri and Choi, Yejin and others},
  journal={arXiv preprint arXiv:2401.03568},
  year={2024}
}

@article{xi2025rise,
  title={The rise and potential of large language model based agents: A survey},
  author={Xi, Zhiheng and Chen, Wenxiang and Guo, Xin and He, Wei and Ding, Yiwen and Hong, Boyang and Zhang, Ming and Wang, Junzhe and Jin, Senjie and Zhou, Enyu and others},
  journal={Science China Information Sciences},
  volume={68},
  number={2},
  pages={121101},
  year={2025},
  publisher={Springer}
}

@article{cheng2024exploring,
  title={Exploring large language model based intelligent agents: Definitions, methods, and prospects},
  author={Cheng, Yuheng and Zhang, Ceyao and Zhang, Zhengwen and Meng, Xiangrui and Hong, Sirui and Li, Wenhao and Wang, Zihao and Wang, Zekai and Yin, Feng and Zhao, Junhua and others},
  journal={arXiv preprint arXiv:2401.03428},
  year={2024}
}

@article{li2025deepagent,
  title={DeepAgent: A General Reasoning Agent with Scalable Toolsets},
  author={Li, Xiaoxi and Jiao, Wenxiang and Jin, Jiarui and Dong, Guanting and Jin, Jiajie and Wang, Yinuo and Wang, Hao and Zhu, Yutao and Wen, Ji-Rong and Lu, Yuan and others},
  journal={arXiv preprint arXiv:2510.21618},
  year={2025}
}

@article{kanfar2025intelligent,
  title={Intelligent seismic workflows: The power of generative AI and language models},
  author={Kanfar, Rayan and Alali, Abdulmohsen and Tonellot, Thierry-Laurent and Salim, Hussain and Ovcharenko, Oleg},
  journal={The Leading Edge},
  volume={44},
  number={2},
  pages={142--151},
  year={2025},
  publisher={Society of Exploration Geophysicists}
}

@misc{SPECFEM3D_GitHubDevel,
  author       = {{SPECFEM Developers}},
  title        = {SPECFEM3D Cartesian: Seismic Wave Propagation Software (devel branch)},
  howpublished = {\url{https://github.com/SPECFEM/specfem3d}},
  note         = {Accessed: 2025-06-20. Commit from the devel branch.},
  year         = {2025}
}

%
%
%
%
%

\end{document}